\documentclass[conference]{IEEEtran}
\usepackage{times}

\usepackage{tikz}
\usetikzlibrary{shapes}
\usetikzlibrary{positioning, arrows.meta, shapes.geometric, calc}

\usepackage{pgfplots}
  \usepgfplotslibrary{groupplots}                                                                                          
\usepackage[caption=false, font=footnotesize]{subfig}
\usepackage{amsmath}
\usepackage{dsfont}
\usepackage[numbers]{natbib}
\usepackage{multicol}
\usepackage{xr-hyper} 

\usepackage[bookmarks=true]{hyperref}
\usepackage{booktabs}
\usepackage[normalem]{ulem}
\usepackage[table]{xcolor}  
\usepackage{amsfonts}
\usepackage{graphicx} 
\usepackage{multirow}
\usepackage{tikz}
\usepackage{xcolor}
\usetikzlibrary{positioning, fit, backgrounds, calc, shapes.arrows, shadows}

\definecolor{best}{RGB}{190, 240, 190} 
\definecolor{second}{RGB}{250, 250, 190} 
\definecolor{trajColor}{RGB}{229, 137, 50}   
\definecolor{queryColor}{RGB}{76, 133, 196}  
\definecolor{decoColor}{RGB}{106, 168, 79}   
\definecolor{darkgray}{RGB}{80, 80, 80}
\definecolor{scopeGlobal}{RGB}{207, 226, 243} 
\definecolor{scopeFrame}{RGB}{255, 242, 204}  

\newcommand{\csec}[1]{\cellcolor{second}#1}
\newcommand{\methodname}{LoTIS}
\newcommand{\projectpage}{\url{https://finnbusch.com/lotis}}
\newcommand{\res}[2]{#1 \textcolor{gray}{\scriptsize{(#2)}}}
\newcommand{\resb}[2]{\textbf{#1} \textcolor{gray}{\scriptsize{(\textbf{#2})}}}

\newcommand{\cb}[1]{\cellcolor{best}#1}

\pgfplotsset{compat=1.17}
\graphicspath{{figures/}{figures/master_vggt/}} 
\definecolor{finn}{RGB}{255, 0, 0}
\definecolor{matti}{RGB}{0, 0, 255}
\definecolor{jesus}{RGB}{0, 100, 0}
\definecolor{quantao}{RGB}{255, 0, 255}
\definecolor{jana}{RGB}{0, 155, 155}

\newcommand{\startpos}{\tikz[baseline=-0.5ex]{\fill[blue!70] (0,0) circle (2pt);}}
\newcommand{\finalpos}{\tikz[baseline=-0.5ex]{\fill[red!80] (0,0) circle (2pt);}}
\newcommand{\initpos}{\tikz[baseline=-0.5ex]{\fill[green!80] (0,0) circle (2pt);}}

\newcommand{\refview}[1]{\tikz[baseline=(text.base)]\node[fill=orange!20, inner sep=2pt, rounded corners=2pt] (text) {#1};}
\newcommand{\extraview}[1]{\tikz[baseline=(text.base)]\node[fill=blue!15, inner sep=2pt, rounded corners=2pt] (text) {#1};}
\newcommand{\onboardview}[1]{\tikz[baseline=(text.base)]\node[fill=green!15, inner sep=2pt, rounded corners=2pt] (text) {#1};}
\IEEEoverridecommandlockouts
\begin{document}

\title{
Learning to Localize Reference Trajectories in Image-Space for Visual Navigation
}
%
\author{
  \authorblockN{Finn Lukas Busch, Matti Vahs, Quantao Yang, Jesús Gerardo Ortega Peimbert,\\Yixi Cai, Jana Tumova, Olov Andersson
  \thanks{This work was partially supported by the Wallenberg AI, Autonomous Systems and Software Program (WASP) funded by the Knut and Alice Wallenberg Foundation. The development was partly enabled by the Berzelius resource provided by the Knut and Alice Wallenberg Foundation at the National Supercomputer Centre.}
  \thanks{The authors are with the Division of Robotics, Perception, and Learning, KTH Royal Institute of Technology, Sweden, and also affiliated with Digital Futures. Contact: \texttt{\{flbusch, vahs, quantao, jgop, yixica, tumova, olovand\}@kth.se}{\tt\small}}%
}}
%


%
%
%
\maketitle
\begin{abstract}
We present \methodname{}, a model for visual navigation that provides robot-agnostic image-space guidance by localizing a reference RGB trajectory in the robot’s current view, without requiring camera calibration, poses, or robot-specific training.
Instead of predicting actions tied to specific robots, we predict the image-space coordinates of the reference trajectory as they would appear in the robot’s current view. This creates robot-agnostic visual guidance that easily integrates with local planning. Consequently, our model's predictions provide guidance zero-shot across diverse embodiments. By decoupling perception from action and learning to localize trajectory points rather than imitate behavioral priors, we enable a cross-trajectory training strategy that learns robust invariance to viewpoint and camera changes. We outperform state-of-the-art methods by 20-50 percentage points in success rate on forward navigation, and paired with a local planner we achieve 94-98\% success rate across diverse sim and real environments. Furthermore, we achieve over 5$\times$ improvements on challenging tasks where baselines fail, such as backward traversal. The system is lightweight ($\sim 22 \, \mathrm{Hz}$ on Jetson Orin AGX) and straightforward to use: we show how even a video from a handheld phone camera directly enables different robots to navigate to any point on the trajectory. Videos, demo, and code are available at \projectpage.
\end{abstract}

\IEEEpeerreviewmaketitle
\section{Introduction}
%
Visual navigation enables robots to navigate environments using only camera observations. Early work focused on navigating to a single goal image. To enable long range navigation, subsequent work has considered visual reference trajectories, i.e. sequences of unposed RGB images recorded along a path that capture the full route to be followed. Such trajectories can be recorded with any camera, from a handheld smartphone to a robot-mounted sensor, allowing routes to be defined through demonstration without requiring metric maps.


State-of-the-art approaches typically address this task via end-to-end learning, training policies that map current observations and a goal image directly to robot actions~\cite{shah2023vint, sridhar2024nomad, suomela2024placenav}. To follow trajectories, these methods extract a single subgoal image from the trajectory and use this as the goal image for the learned policy. We identify three limitations.
First, outputting actions directly constrains models to the training action space (typically planar differential drive), limiting generalization to platforms like aerial robots.
Second, to learn a policy mapping from observations to actions, these methods require training examples where the robot traverses from the current view to the goal image. This restricts training to start and goal images from the same trajectory, and thus implicitly assumes that the deployment camera shares the characteristics of the recording device. We show that deviations in camera intrinsics or robot embodiment therefore lead to degraded performance.
Third, relying on a single subgoal image makes the system sensitive to subgoal selection errors. If an incorrect target is chosen, the navigation policy may fail, as the robot cannot visually match its current view to the incorrect subgoal.
\begin{figure}[t]
\centering
\def\svgwidth{\columnwidth}
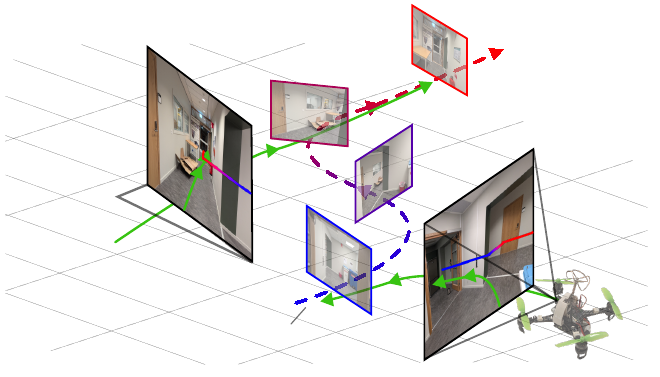
\vspace{-0.6cm}
\caption{Given only a reference trajectory of (unposed) RGB images $\mathcal{T}$, our model localizes the trajectory within the robot's current view. The predicted image-space coordinates, distances and visibility of the reference trajectory poses $(\mathbf{p}_i, d_i, v_i)$ act as an interface for local planning, allowing any robot such as quadrupeds or drones to go to any point on the trajectory, from any view of the trajectory.}
\label{fig:figure1}
\end{figure}

We approach the problem of visual navigation by decoupling perception from action, and propose to learn a model that predicts the reference trajectory directly in image space, such that it can be tracked by conventional local planners.

To this end, we present \methodname{}, a model for visual navigation that given a reference trajectory (sequence of RGB images) predicts the following representation: 1) the image-space coordinates of where each trajectory pose would appear in the robot's current view, 2) whether these poses are visible, and 3) a normalized distance to each visible pose, see Fig.~\ref{fig:figure1}. This overcomes the aforementioned challenges:
First, our model predictions can be easily paired with downstream controllers or planners for diverse robot embodiments, without further training. Second,
because our model learns to localize trajectory poses rather than imitate actions, we can construct training pairs by sampling reference and query images from different trajectories. As a result, we can explicitly train on data from mismatched cameras, varied mounting heights, and off-trajectory viewpoints, enabling robustness to such variations.
Finally, processing the full trajectory sequence rather than selecting a single subgoal image overcomes the sensitivity of accurate subgoal selection in existing works, leading to substantial performance gains in navigation success rate across all evaluations.

In addition, our model enables new capabilities, most notably backward traversal, where related approaches largely fail while our method maintains robust performance. This enables, for the first time, using an RGB reference trajectory in the general setting where the robot can navigate from anywhere in the vicinity (with visual overlap) to any point on the trajectory, both forward and backward. 

In summary, we decouple perception from action and provide a learned perception model that interfaces with classical planners, with the following contributions:
\begin{enumerate}
    \item We propose an image-space representation, defined by 2D coordinates, visibility, and distance of each reference trajectory pose in the robot's current view, that easily pairs with local planners to guide diverse robots in image space. In real-world experiments, we show that this representation is suitable to guide both a quadrotor and a quadruped from a single phone-recorded trajectory.
    \item We introduce a cross-trajectory training strategy tailored for this representation, where reference and query images are sampled from different trajectories, exposing the model to camera mismatches and challenging viewpoints. This enables robust backward traversal and consequently supports navigation to any point on the trajectory.
    \item We propose a model architecture that processes the full reference trajectory jointly rather than relying on single subgoal-selection, while enabling real-time deployment on embedded hardware. We show that for existing methods, reliance on a single subgoal-selection leads to decreased localization accuracy with distance from the trajectory, whereas our joint processing enables \methodname{} paired with a local planner to maintain robust success rates even when initialized far from the trajectory.
\end{enumerate}
\section{Related Work}
\label{sec:related_work}
\subsection{Visual Navigation}
The core challenge of visual navigation is translating image observations into physical motion, a task traditionally addressed by visual servoing. Visual servoing stands as the classical foundation for visual navigation, providing a framework for translating sensor visual feedback into control actions, with Image-Based Visual Servoing (IBVS) minimizing error directly in feature space and Position-Based Visual Servoing (PBVS) operating through explicit pose estimation~\cite{espiau1991new, chaumette2006visual, chaumette2007visual}. However, these methods suffer from well-documented limitations: local minima, sensitivity to calibration errors, and reliance on continuous feature tracking that breaks under occlusions or appearance changes~\cite{chaumette1998potential}. These challenges motivated learning-based approaches that can extract features robust to appearance variation and operate without explicit calibration.
\begin{figure*}[t]
\centering
\def\svgwidth{\textwidth}
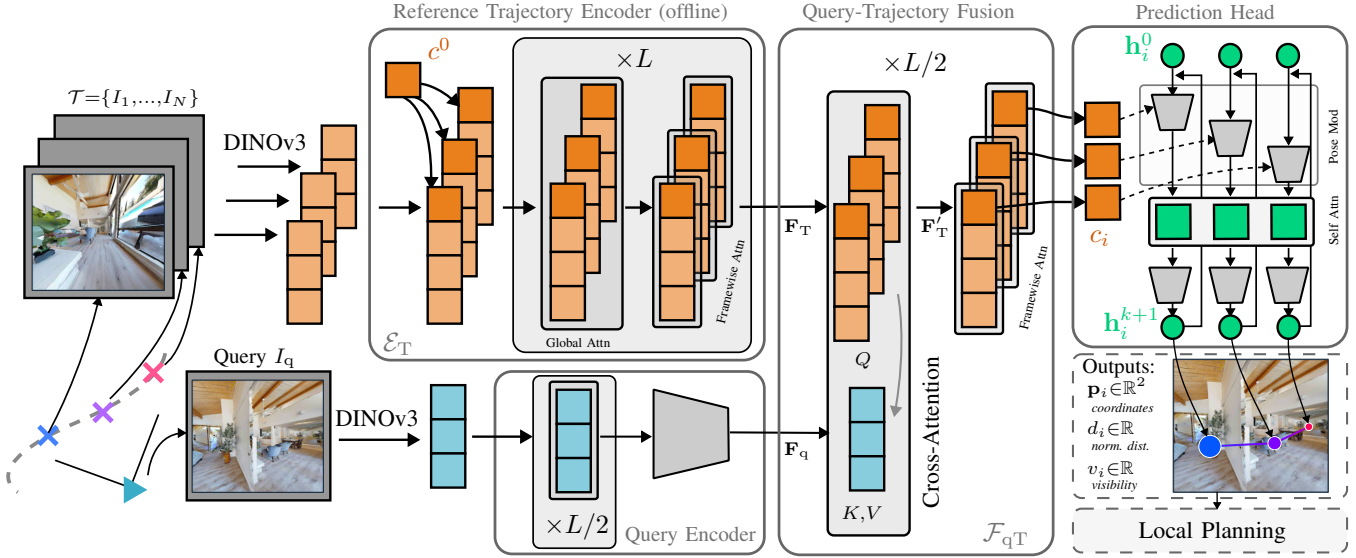
\vspace{-0.6cm}
\caption{\textbf{\methodname{} Architecture.} Reference trajectory $\mathcal{T}$ and query $I_q$ are processed by frozen DINOv3 backbones. A trajectory encoder ($\mathcal{E}_T$) captures spatio-temporal context once (offline), while a query encoder ($\mathcal{E}_q$) and query-trajectory fusion ($\mathcal{F}_{qT}$) perform online feature extraction and fusion, respectively. Finally, a recurrent transformer iteratively regresses image-space coordinates ($\mathbf{p}_i$), visibility ($v_i$), and distances ($d_i$).}
\vspace{-0.4cm}
\label{fig:architecture}
\end{figure*}

Modern learning-based visual navigation has evolved from image-goal navigation benchmarks~\cite{krantz2022instance, krantz2023navigating} toward topological methods designed for real-world deployment. GNM~\cite{shah2023gnm} introduced a cross-embodiment navigation policy trained on heterogeneous robot data, predicting both temporal distance for subgoal selection and normalized waypoint actions. ViNT~\cite{shah2023vint} scaled this approach using a Transformer architecture trained on over one hundred hours of diverse navigation data. NoMaD~\cite{sridhar2024nomad} unified goal-directed navigation and exploration within a diffusion policy, using goal masking to enable flexible inference.

Since performance largely depends on accurate subgoal selection, PlaceNav~\cite{suomela2024placenav} aims to improve robustness by reframing subgoal selection as visual place recognition using CosPlace~\cite{berton2022cosplace}, applying Bayesian filtering for temporal consistency while relying on GNM for low-level waypoint control. FAINT~\cite{suomela2025faint} combines EigenPlaces~\cite{berton2023eigenplaces} for place recognition with a navigation policy trained on frozen Theia~\cite{shang2025theia} representations, demonstrating that simulation-trained policies can outperform real-world-trained counterparts when sufficient synthetic data is available.

These methods share fundamental limitations that our work addresses. First, they couple perception with learned behavioral priors, outputting actions tied to specific robot kinematics. While these approaches train on data from multiple robots, they remain constrained to platforms with similar action spaces (e.g., ground-based differential drive), limiting generalization to platforms with different motion capabilities such as aerial robots. Second, they are primarily designed for forward trajectory following and struggle with backward traversal, where the robot encounters viewpoints substantially different from those in the recorded trajectory. Third, they are sensitive to camera mismatch between recording and deployment, as training on on-trajectory data does not expose the models to such variations.

Concurrent image-space local navigation methods such as VAMOS~\cite{castro2025vamos} and VENTURA~\cite{zhang2025ventura} also aim to improve cross-embodiment generalization by predicting local traversability or local paths from the current image. However, these methods address a complementary problem: local navigation toward a nearby language or image goal from the current observation. In contrast, LoTIS addresses reference-trajectory navigation, where the robot must localize itself relative to a pre-recorded image sequence and navigate to arbitrary trajectory indices, including distant goals, multi-turn routes, and backward traversal. Thus, local image-space planners could in principle be combined with LoTIS as downstream local controllers, while LoTIS provides the long-range trajectory localization signal.
\subsection{Learned Visual Geometry}
The field of learned visual geometry is related to visual navigation through a shared need for spatial understanding from images. In this domain, recent models achieve remarkable geometric understanding from images. DUSt3R~\cite{wang2024dust3r} recovers dense pointmaps from image pairs without calibration, MASt3R~\cite{leroy2024mast3r} augments this with dense local features and a matching loss for robust correspondences under extreme viewpoint changes, VGGT~\cite{wang2025vggt} extends to joint pose and geometry estimation, and Depth~Anything~3~\cite{lin2025depth} unifies multi-view depth and pose estimation through a depth-ray representation with a plain DINOv2 backbone.

These methods are primarily designed for 3D reconstruction and mapping rather than online robot control, and have not yet been adapted into real-time navigation policies. We found that applying these state-of-the-art reconstruction models to our evaluation trajectories (see Fig.~\ref{fig:realworld_setup}) often yielded inconsistent poses or failed reconstructions. We suspect that solving the full 3D reconstruction problem, which prioritizes global metric accuracy, is computationally heavier and likely harder than the task of relative trajectory localization required for navigation.  We provide examples covering all of our real-world evaluation trajectories in Appendix~\ref{s_sec:vggt_study}. \methodname{} addresses this by learning a representation explicitly tailored for navigation. Instead of recovering general scene geometry, our model is designed to provide the specific information required for navigation: the location of the reference trajectory relative to the robot’s current view.

\section{Problem Statement}
\label{sec:problem}
We consider the task of navigating to any point along a recorded RGB reference trajectory using only RGB images. Formally, given a reference trajectory $\mathcal{T} = \{I_1, \ldots, I_N\}$ consisting of $N$ RGB images, and a query image $I_\mathrm{q}$ from the robot's current viewpoint, the goal is to navigate to any point on the trajectory, indexed by $g \in \{1, \dots, N\}$, starting from any position in the trajectory's vicinity, as long as there is visual overlap with some portion of $\mathcal{T}$.

We make no explicit assumptions about camera calibration or robot embodiment. As such the reference trajectory $\mathcal{T}$ may be recorded with a different camera or platform than the one used for navigation.
\section{Method}
We approach this problem by decoupling perception from action and learn a model that provides robot-agnostic guidance that can be easily consumed by classical motion planning and control stacks designed for specific embodiments. 
To this end, we propose \methodname{}, a model that predicts where the poses of a reference trajectory appear in the robot’s current view in image-space. As this output is formulated in the robot's current view, it can easily be used by downstream controllers, e.g. by simply steering towards the predicted points.

\subsection{Guidance Representation}
\label{sec:representation}
Our model's output provides guidance to be used by classical motion planning stacks.
Formally, in the image-space of the robot's current view, we predict a triplet $(\mathbf{p}_i, v_i, d_i)$ for each reference frame $I_i \in \mathcal{T}$:
\begin{itemize}
    \item A 2D point in image-space $\mathbf{p}_i \in \mathbb{R}^2$: where the camera pose corresponding to trajectory image $I_i$ would appear in the query view, normalized to $[-1, 1]\times [-1,1]$
    \item A visibility logit $v_i \in \mathbb{R}$: whether the pose is visible or occluded/out of view
    \item A normalized distance $d_i \in [0, 1]$: relative distance to the pose from the current query viewpoint. For scale-independency, we normalize the distances such that the farthest visible point is at distance 1.
\end{itemize}
\subsection{Model Architecture}
\label{sec:architecture}
We design our model to process the reference trajectory $\mathcal{T}$ and match this with the current view of the robot $I_\mathrm{q}$. Unlike most existing methods~\cite{shah2023vint, sridhar2024nomad, suomela2025faint, suomela2024placenav} that match the current view pairwise to each image on the trajectory, we propose processing the full trajectory jointly and matching the robot's view against that. To ensure real-time deployment, we propose an asymmetric model architecture: a large \emph{trajectory encoder} $\mathcal{E}_\mathrm{T}$ runs once at deployment to process the full trajectory, while a lightweight \emph{query encoder} $\mathcal{E}_\mathrm{q}$ and \emph{decoder} run efficiently online to produce the final predictions. The decoder consists of query-trajectory fusion $\mathcal{F}_\mathrm{qT}$, which matches $I_\mathrm{q}$ against the processed trajectory to find visual correspondences, and aggregates those into a global context within one summary token $\mathbf{c}_i$ per frame $i$ on the reference trajectory; and a  prediction head predicts the one final prediction $\mathbf{p}_i, v_i, d_i$ per frame $i$ from the summary tokens. An overview is provided in Fig.~\ref{fig:architecture}.

%
We use DINOv3~\cite{simeoni2025dinov3} as a frozen backbone to extract initial features for all images, given the strong performance of the DINO family on geometric vision tasks~\cite{wang2025vggt}. All images are resized to $224 \times 224$ before DINOv3.
\subsubsection{Trajectory Encoder $\mathcal{E}_\mathrm{T}$}
Given the reference trajectory $\mathcal{T} = \{I_1, \ldots, I_N\}$, we extract per-frame features using frozen DINOv3, project to dimension $D$, and prepend a learnable \emph{summary token} $\mathbf{c}_0$ per frame. The resulting tokens are processed by $L$ transformer blocks, each alternating between global attention (across all frames and patches) and frame-wise attention (within each frame), following~\cite{wang2025vggt}. We incorporate temporal structure via rotary position encodings (RoPE~\cite{su2024rope}). The encoder outputs $\mathbf{F}_\mathrm{T} \in \mathbb{R}^{N \times (P+1) \times D}$, where $P$ is the number of patches per image.
\subsubsection{Query Encoder $\mathcal{E}_\mathrm{q}$}
The query image $I_q$ is processed by the same frozen DINOv3 backbone, projected to dimension $D$, and refined through $L/2$ self-attention layers with spatial RoPE, producing tokens $\mathbf{F}_\mathrm{q} \in \mathbb{R}^{P \times D}$. A linear adapter aligns features with the trajectory encoder's representation space.
\subsubsection{Query-Trajectory Fusion $\mathcal{F}_\mathrm{qT}$}
This module fuses query and trajectory features $\mathbf{F}_\mathrm{q}, \mathbf{F}_\mathrm{T}$ through $L/2$ blocks, each alternating cross-attention and frame-wise self-attention. For cross-attention, trajectory features $\mathbf{F}_\mathrm{T}$ (patches and summary tokens) serve as queries, while query image tokens $\mathbf{F}_\mathrm{q}$ serve as keys and values:
\begin{equation}
    \mathbf{F}_\mathrm{T}' = \mathrm{CrossAttn}(Q{=}\mathbf{F}_\mathrm{T}, K{=}\mathbf{F}_\mathrm{q}, V{=}\mathbf{F}_\mathrm{q}).
\end{equation}
This allows each trajectory frame to attend to the query view and find visual correspondences. The subsequent frame-wise self-attention aggregates these local correspondences into global context within each frame's summary token $\mathbf{c}_i$.
\subsubsection{Prediction Head $\mathcal{P}$}
We employ a recurrent transformer regressor, adapting the iterative refinement architecture from~\cite{wang2025vggt}. The head maintains a latent estimate $\mathbf{h}_i \in \mathbb{R}^{D}$ for each trajectory frame $i \in \{1, \ldots, N\}$, initialized by a learnable query. Over $K$ iterations, the current estimate $\mathbf{h}^{(k)}$ is projected and used to condition the encoded summary tokens $\mathbf{c}_i$ via Adaptive Layer Normalization (AdaLN~\cite{peebles2023scalable}). The modulated tokens are processed by self-attention across all $N$ frames, enabling the model to enforce geometric consistency along the trajectory. A linear layer predicts residual updates, and the estimate is refined as $\mathbf{h}^{(k+1)} \leftarrow \mathbf{h}^{(k)} + \Delta \mathbf{h}$.

After $K$ iterations, the final estimates are projected to the output space: image coordinates $\mathbf{p}_i$ via $\tanh$ (normalized to $[-1,1]^2$), visibility logits $v_i$, and normalized distances $d_i$ via scaled $\tanh$ (bounded to $[0,1]$).
\subsubsection{Implementation}
We use $L{=}12$ layers, hidden dimension $D{=}256$, and $K{=}4$ refinement iterations. This yields ${\sim}50\,\mathrm{M}$ parameters, with ${\sim}32\,\mathrm{M}$ in the trajectory encoder. Further details on the model and scaling are provided in Appendix~\ref{s_sec:impl_details}.
\subsection{Training}
\label{sec:training}
%
%
\subsubsection{Cross-Trajectory Training}
We train on a combination of real-world navigation datasets and synthetic data generated in simulation~\cite{habitat19iccv}.

A key advantage of our representation is that it enables \emph{cross-trajectory} sampling: For real-world datasets, we sample the reference trajectory $\mathcal{T}$ and query image $I_\mathrm{q}$ from \emph{different} trajectories within the same environment. We do this to encourage generalization, as this exposes the model to camera mismatches, off-trajectory viewpoints and environmental variations between traversals (lighting changes, dynamic objects, seasonal differences).
Importantly, cross-trajectory sampling is only possible because we do not require action annotations between reference trajectory and query image. 

In simulation, we generate reference trajectories between random start and goal points, sampling query images from random poses in the vicinity of these trajectories. To encourage generalization, we independently randomize camera parameters (field-of-view, aspect ratio, mounting height) for reference and query views, and apply rotational perturbations to query and trajectory poses before recording the images. Trajectory frames are sampled at stochastic intervals to vary sequence density for both simulation and real-world datasets.

Ground-truth image coordinates and distances are obtained by projecting trajectory camera centers into the query camera using camera poses. Visibility labels are generated by checking whether the projected point lies inside the image and is not occluded; depth maps are used only for this occlusion check. For real-world data, we preprocess depth with PriorDepthAnything~\cite{wang2025priorDA}.
%
%
%
%
\subsubsection{Losses}
We optimize the model using a weighted sum of three objectives:
\begin{equation}
    \label{eq:loss}
    \mathcal{L} = \lambda_{\mathrm{pos}} \mathcal{L}_{\mathrm{pos}} + \lambda_{\mathrm{vis}} \mathcal{L}_{\mathrm{vis}} + \lambda_{\mathrm{dist}} \mathcal{L}_{\mathrm{dist}},
\end{equation}
where $\mathcal{L}_{\mathrm{pos}}$ is an $L_1$ loss applied to the predicted coordinates $\mathbf{p}_i$, masked to only penalize points where the ground truth is visible ($v^*_i=1$). 
$\mathcal{L}_{\mathrm{vis}}$ is a Binary Cross-Entropy loss applied to the visibility logits.
$\mathcal{L}_{\mathrm{dist}}$ is an $L_1$ loss applied to the normalized distance predictions. We compute this loss on the output of each iteration of the prediction head and apply temporal weighting~\cite{teed2020raft}.
\subsubsection{Datasets}
Our training data spans diverse environments, including simulation (HM3D~\cite{ramakrishnan2021hm3d}, HSSD~\cite{khanna2024hssd}, AI2-THOR~\cite{kolve2017ai2}) and real-world trajectories (CODa~\cite{zhang2024coda}, LILocBench~\cite{trekel2025bonn}, BotanicGarden~\cite{liu2024botanicgarden}, TartanGround~\cite{patel2025tartanground}). This mixture exposes the model to cluttered indoor spaces, unstructured outdoor paths, and urban settings with dynamic objects, as well as seasonal and day-night variations.

In total, we use approximately 25,000 reference trajectories (up to 40 frames each) and 850,000 query images across 500 unique environments. We train on a single NVIDIA RTX 5090 for approximately 4 days.
\subsection{Navigation}
\label{sec:control}
Our model outputs a set of points $\mathbf{p}_i$ in the robot's current image frame representing the reference trajectory. To navigate, a downstream controller uses these predictions alongside a user-specified goal index $g \in \{1, \dots, N\}$ to determine the appropriate actions. We demonstrate this flexibility on two distinct controllers:

\noindent \textbf{1) Yaw Controller with constant forward velocity:} To demonstrate robust performance with minimal complexity, we employ a simple controller that controls only yaw velocity while commanding constant forward velocity. The controller identifies visible points $\mathcal{I}_{\mathrm{vis}} = \{ i \mid \sigma(v_i) > 0.5 \}$, selects the closest one ($k = \operatorname*{argmin}_{i \in \mathcal{I}_{\mathrm{vis}}} d_i$), and applies an offset in the desired direction of travel. We apply a proportional controller to derive yaw velocity commands that steer toward this target while commanding constant forward velocity.

%
\noindent \textbf{2) Model Predictive Path Integral Control:}
To demonstrate that our model predictions can be easily combined with more sophisticated control strategies, we employ a perception-aware Model Predictive Path Integral (MPPI) controller~\cite{williams2016mppi} for a drone. We formulate a cost function that ensures that our predictions remain in the robot's FOV, aligning with similar approaches in perception-aware MPC~\cite{falanga2018pampc, mohamed2021sampling}, and further incorporate proactive collision-avoidance. To this end, we use depth maps predicted by UniDepthV2~\cite{piccinelli2025unidepthv2} to ground the model's predictions in 3D and to formulate collision avoidance.
We refer the reader to Appendix~\ref{s_sec:contr_details} for details on both controllers.
\section{Simulation Experiments}
\label{sec:sim_experiments}

\begin{table*}[t]
\centering
\setlength{\tabcolsep}{3.5pt} 
\renewcommand{\arraystretch}{1.15}
\caption{\textbf{Navigation Performance}: We report Success Rate (SR) and SPL (in parentheses). \textbf{Cross}: Query camera differs from reference camera. \textbf{Matched}: Same camera parameters. \textbf{Off-Trajectory}: Robot initialized away from the Trajectory. \textbf{On-Trajectory}: Robot initialized on the Trajectory. For details, see Sec.~\ref{subsec:sim_setup}.}
\vspace{-0.3cm}
\label{tab:unified_results}
\begin{tabular}{c l cc cc | cc cc | cc }
\toprule
 & & \multicolumn{4}{c|}{\textbf{To End (Forward)}} & \multicolumn{4}{c|}{\textbf{To Start (Backward)}} & \multicolumn{2}{c}{\textbf{Any Point}} \\
 & & \multicolumn{2}{c}{On-Trajectory} & \multicolumn{2}{c|}{Off-Trajectory} & \multicolumn{2}{c}{On-Trajectory} & \multicolumn{2}{c|}{Off-Trajectory} & \multicolumn{2}{c}{Off-Trajectory} \\
\cmidrule(lr){3-4} \cmidrule(lr){5-6} \cmidrule(lr){7-8} \cmidrule(lr){9-10} \cmidrule(lr){11-12}
 & \textbf{Method} & Matched & Cross & Matched & Cross & Matched & Cross & Matched & Cross & Matched & Cross \\
\midrule

\multirow{6}{*}{\rotatebox[origin=c]{90}{\textbf{Gibson}}} & ViNT~\cite{shah2023vint} & \csec{\res{70.4}{67.8}} & \res{21.1}{17.8} & \res{40.1}{30.7} & \res{21.1}{17.5} & \res{1.3}{0.9} & \res{3.9}{3.4} & \res{9.2}{8.1} & \res{9.9}{7.8} & \res{27.6}{22.8} & \res{21.7}{19.3} \\
 & PlaceNav~\cite{suomela2024placenav} & \res{53.9}{52.0} & \res{5.3}{4.9} & \res{15.1}{12.8} & \res{11.2}{10.5} & \res{3.9}{3.7} & \res{4.6}{3.9} & \res{9.2}{9.0} & \res{10.5}{9.9} & \res{22.4}{19.5} & \res{13.8}{12.2} \\
 & NoMaD~\cite{sridhar2024nomad} & \res{23.0}{20.2} & \res{8.6}{7.4} & \res{24.3}{17.0} & \res{17.1}{11.8} & \res{8.6}{7.0} & \res{7.2}{5.8} & \res{11.2}{8.7} & \res{11.2}{9.0} & \res{31.6}{22.4} & \res{27.0}{19.8} \\
 & FAINT~\cite{suomela2025faint} & \res{50.7}{47.8} & \csec{\res{34.2}{31.0}} & \csec{\res{50.0}{42.5}} & \csec{\res{41.4}{33.6}} & \csec{\res{11.8}{9.3}} & \csec{\res{9.2}{7.5}} & \csec{\res{11.2}{10.1}} & \csec{\res{13.2}{10.4}} & \csec{\res{52.0}{42.6}} & \csec{\res{40.8}{35.2}} \\
 & \methodname{} (Ours) & \cb{\resb{94.7}{94.7}} & \cb{\resb{83.6}{83.1}} & \cb{\resb{88.2}{85.0}} & \cb{\resb{82.2}{79.5}} & \cb{\resb{88.2}{84.2}} & \cb{\resb{80.9}{78.1}} & \cb{\resb{77.6}{72.4}} & \cb{\resb{71.7}{67.7}} & \cb{\resb{85.5}{82.7}} & \cb{\resb{81.6}{77.8}} \\
  & + Obstcl. Avoidance & \cb{\resb{100.0}{99.9}} & \cb{\resb{98.0}{96.1}} & \cb{\resb{98.7}{93.9}} & \cb{\resb{94.7}{87.8}} & \cb{\resb{97.4}{89.9}} & \cb{\resb{89.5}{83.6}} & \cb{\resb{96.7}{87.6}} & \cb{\resb{90.8}{81.9}} & \cb{\resb{98.0}{92.4}} & \cb{\resb{95.4}{87.9}} \\
\midrule

\multirow{6}{*}{\rotatebox[origin=c]{90}{\textbf{HM3D}}} & ViNT~\cite{shah2023vint} & \csec{\res{65.7}{64.4}} & \res{12.7}{12.1} & \res{24.0}{20.8} & \res{12.3}{10.3} & \res{3.4}{3.1} & \res{2.5}{2.3} & \res{4.4}{3.7} & \res{5.4}{5.0} & \res{20.1}{17.9} & \res{10.3}{9.3} \\
 & PlaceNav~\cite{suomela2024placenav} & \res{55.4}{54.5} & \res{7.8}{7.3} & \res{11.3}{10.0} & \res{6.4}{5.0} & \res{3.4}{3.2} & \res{2.5}{2.1} & \res{4.9}{3.4} & \res{4.9}{4.0} & \res{13.2}{11.4} & \res{8.3}{7.5} \\
 & NoMaD~\cite{sridhar2024nomad} & \res{25.0}{22.2} & \res{9.3}{8.2} & \res{14.2}{11.2} & \res{12.7}{10.0} & \res{4.4}{4.0} & \res{4.4}{3.9} & \res{10.3}{8.9} & \res{8.3}{7.1} & \res{23.5}{16.3} & \res{14.7}{12.3} \\
 & FAINT~\cite{suomela2025faint} & \res{60.3}{59.0} & \csec{\res{34.8}{31.6}} & \csec{\res{46.1}{40.6}} & \csec{\res{28.9}{25.3}} & \csec{\res{7.8}{7.1}} & \csec{\res{11.3}{10.3}} & \csec{\res{10.8}{9.6}} & \csec{\res{10.8}{9.3}} & \csec{\res{36.3}{32.4}} & \csec{\res{26.5}{23.5}} \\
 & \methodname{} (Ours) & \cb{\resb{98.5}{98.4}} & \cb{\resb{90.2}{90.1}} & \cb{\resb{74.0}{72.3}} & \cb{\resb{74.5}{73.0}} & \cb{\resb{81.9}{79.4}} & \cb{\resb{69.6}{67.6}} & \cb{\resb{69.6}{65.6}} & \cb{\resb{65.2}{61.6}} & \cb{\resb{71.1}{69.5}} & \cb{\resb{71.6}{70.4}} \\
  & + Obstcl. Avoidance  & \cb{\resb{100.0}{99.8}} & \cb{\resb{97.5}{96.7}} & \cb{\resb{95.1}{89.6}} & \cb{\resb{92.2}{85.1}} & \cb{\resb{96.1}{90.7}} & \cb{\resb{84.8}{80.2}} & \cb{\resb{92.6}{83.2}} & \cb{\resb{86.8}{76.2}} & \cb{\resb{94.1}{90.0}} & \cb{\resb{94.1}{88.2}} \\
\bottomrule
\end{tabular}
\vspace{-0.4cm}
\end{table*}
We evaluate our method in photo-realistic simulation to assess robustness to environmental variations and to enable reproducible experiments, addressing the following questions:
\begin{itemize}
    \item[\textbf{Q1:}] How well does our method perform on forward trajectory following compared to baselines that were specifically designed for this task?    
    \item[\textbf{Q2:}] How does our model handle off-trajectory starts compared to methods that rely on discrete subgoal retrieval?
    \item[\textbf{Q3}:] How robust is our method to mismatched camera intrinsics and mounting heights between reference and query trajectories?
    \item[\textbf{Q4:}] To what extent does a model trained only on forward trajectories generalize to backward traversal without explicit backward traversal demonstrations?
\end{itemize}

\subsection{Experimental Setup}
\label{subsec:sim_setup}
We utilize the Gibson Habitat split~\cite{xiazamirhe2018gibsonenv} (5 environments) and HM3D~\cite{ramakrishnan2021hm3d} (100 environments) datasets within the Habitat simulator. All evaluation environments are held out from training. Hyperparameters were tuned during initial development on training environments and then fixed across all reported evaluations, baseline parameters were tuned for best performance. Overall, we evaluate on 100 reference trajectories for each dataset using two randomized initial poses for the robot per trajectory for each evaluation setup.
To test robustness against camera mismatch, we define two evaluation setups: \textit{1) Matched Camera:} The query agent is equipped with the exact camera with same mounting height as the reference trajectory, and \textit{2) Cross-Camera:} The query agent has mismatched FOV (avg.\ $20^\circ$, max $60^\circ$ difference), aspect ratio (avg.\ $0.5$, max $1.5$), and mounting height (avg.\ $0.5\,\mathrm{m}$, max $1.2\,\mathrm{m}$) compared to the recorder. We provide details about the evaluation dataset statistics in Appendix~\ref{s_sec:stats}. Furthermore, we apply rotational noise to the reference trajectory poses before capturing the frames to simulate imperfect recording. We investigate the following three navigation tasks:
\begin{enumerate}
    \item \textbf{To End (Forward Navigation):} The goal is the final frame $I_N$ of the reference trajectory. This is the standard task for the baselines we compare against.
    \item \textbf{To Start (Backward Navigation):} The goal is the first frame $I_1$ of the reference trajectory. The agent must retrace the trajectory in reverse order, i.e. while opposing the views the reference trajectory was recorded from.
    \item \textbf{Any Point (Random Goal):} The goal is a randomly selected frame $I_k$ along the trajectory, requiring the agent to navigate to arbitrary targets within the trajectory.
\end{enumerate}

For the \textbf{To End} and \textbf{To Start} tasks, we evaluate two initialization conditions: (1) \textbf{On-Trajectory:} The agent starts at a pose exactly belonging to the reference trajectory. (2) \textbf{Off-Trajectory:} The agent starts at a random pose in the vicinity of the trajectory with visual overlap. The \textbf{Any Point} task is evaluated only in the off-trajectory setting, as it is intended to assess the most versatile setting, where the agent starts from arbitrary starting positions and is tasked to navigate to any point on the reference trajectory.

\textbf{Metrics.} We report Success Rate (SR) and Success weighted by Path Length (SPL). A run is considered successful if the agent arrives within $0.5\,\mathrm{m}$ of the goal. We terminate a run after 1000 simulation steps.

\textbf{Baselines.} We compare against four state-of-the-art navigation methods using official pre-trained weights. All baselines operate on a topological graph constructed from the reference trajectory: \textbf{ViNT}~\cite{shah2023vint} and \textbf{NoMaD}~\cite{sridhar2024nomad} select subgoals via learned temporal distance, with NoMaD employing a diffusion policy for multimodal action distributions. \textbf{PlaceNav}~\cite{suomela2024placenav} utilizes visual place recognition (CosPlace~\cite{berton2022cosplace}) for retrieval and GNM~\cite{shah2023gnm} for waypoint control. Finally, \textbf{FAINT}~\cite{suomela2025faint} uses EigenPlaces~\cite{berton2023eigenplaces} for retrieval and learns a navigation policy over frozen Theia~\cite{shang2025theia} representations to enhance sim-to-real transfer.

We use the yaw controller with constant forward velocity, see Sec.~\ref{sec:control}, paired with \methodname{} for our simulation experiments. We also report \methodname{} and the controller paired with simple reactive obstacle avoidance to correct movements to ensure collision-free movement. Baselines integrate collision handling into their learned policies and cannot easily be augmented with external modules, illustrating a practical benefit of decoupling perception and control.
\subsection{Results}
We summarize our simulation results in Table~\ref{tab:unified_results}.
\subsubsection{Forward On-Trajectory Navigation}
We first evaluate the standard navigation task: following a reference trajectory forward from an on-trajectory start. \methodname{} achieves a 94.7\% success rate (SR) on Gibson and 98.5\% on HM3D, outperforming the strongest baseline (ViNT) by 24.3 and 32.8 percentage points, respectively.
Notably, when paired with simple obstacle avoidance, our method achieves 100\% SR on both datasets.
We hypothesize that this performance gap is largely due to how the reference trajectory is processed: while baselines extract a single subgoal, \methodname{} localizes the entire sequence in image-space. This likely provides a much richer and more consistent guidance signal, which even a basic downstream controller (controlling only yaw angle) can exploit to achieve superior performance.
%

%
\subsubsection{Off-Trajectory Initialization}
\begin{figure}[t]
    \centering
    \definecolor{myblue}{HTML}{0173B2}
\definecolor{myorange}{HTML}{DE8F05}
\definecolor{myteal}{HTML}{029E73}
\definecolor{mypurple}{HTML}{CC78BC}
\definecolor{myred}{HTML}{D55E00}
\begin{tikzpicture}
  \begin{groupplot}[
      group style={
          group size=2 by 1,
          horizontal sep=0.8cm,             
      },
      width=0.39\columnwidth,  
      height=3cm,
      grid=major,
      grid style={dashed, gray!30},
      ymin=0, ymax=110,
        xmin=1,
      xmax=11,
            enlarge x limits=false,  
      enlarge y limits=false,
      scale only axis,
  ] 
\nextgroupplot[
  xlabel={Init. Distance (m)},
  ylabel={\small Relative SR},
    legend to name=sharedlegend,
    legend columns=5,
legend style={                                                                                                                                                                                                                                                 
      font=\small,                                                                                                                                                                                                                                               
      draw=black,  
      column sep=0.05cm,  
      legend image post style={scale=0.5},  
  },   
]     
\addplot[myblue, thick, smooth] coordinates {(1.505,100.00) (1.822,99.65) (2.140,99.26) (2.457,98.82) (2.775,98.35) (3.092,97.82) (3.410,97.25) (3.727,96.61) (4.045,95.91) (4.362,95.15) (4.680,94.31) (4.998,93.39) (5.315,92.39) (5.633,91.30) (5.950,90.12) (6.268,88.84) (6.585,87.45) (6.903,85.96) (7.220,84.36) (7.538,82.64) (7.855,80.81) (8.173,78.86) (8.490,76.80) (8.808,74.63) (9.125,72.35) (9.443,69.98) (9.760,67.50) (10.078,64.95) (10.395,62.32) (10.713,59.64) (11.031,56.91) (11.348,54.14) (11.666,51.37) (11.983,48.59) (12.301,45.83) (12.618,43.11) (12.936,40.43) (13.253,37.82) (13.571,35.28) (13.888,32.82) (14.206,30.46) (14.523,28.20) (14.841,26.05) (15.158,24.01) (15.476,22.08) (15.793,20.27) (16.111,18.57) (16.429,16.99) (16.746,15.52) (17.064,14.15)};
\addlegendentry{\methodname{}}
\addplot[myorange, thick, smooth] coordinates {(1.505,100.00) (1.822,96.73) (2.140,93.52) (2.457,90.38) (2.775,87.29) (3.092,84.28) (3.410,81.33) (3.727,78.45) (4.045,75.64) (4.362,72.90) (4.680,70.23) (4.998,67.63) (5.315,65.10) (5.633,62.64) (5.950,60.25) (6.268,57.93) (6.585,55.69) (6.903,53.51) (7.220,51.40) (7.538,49.36) (7.855,47.38) (8.173,45.47) (8.490,43.62) (8.808,41.84) (9.125,40.12) (9.443,38.46) (9.760,36.86) (10.078,35.32) (10.395,33.84) (10.713,32.41) (11.031,31.03) (11.348,29.71) (11.666,28.43) (11.983,27.21) (12.301,26.03) (12.618,24.90) (12.936,23.82) (13.253,22.78) (13.571,21.78) (13.888,20.82) (14.206,19.90) (14.523,19.02) (14.841,18.18) (15.158,17.37) (15.476,16.60) (15.793,15.85) (16.111,15.14) (16.429,14.46) (16.746,13.81) (17.064,13.19)};
\addlegendentry{FAINT}
\addplot[myteal, thick, smooth] coordinates {(1.505,100.00) (1.822,92.68) (2.140,85.78) (2.457,79.30) (2.775,73.22) (3.092,67.54) (3.410,62.23) (3.727,57.29) (4.045,52.69) (4.362,48.42) (4.680,44.47) (4.998,40.81) (5.315,37.42) (5.633,34.30) (5.950,31.42) (6.268,28.76) (6.585,26.32) (6.903,24.08) (7.220,22.02) (7.538,20.12) (7.855,18.39) (8.173,16.80) (8.490,15.34) (8.808,14.00) (9.125,12.78) (9.443,11.66) (9.760,10.64) (10.078,9.71) (10.395,8.85) (10.713,8.07) (11.031,7.36) (11.348,6.71) (11.666,6.12) (11.983,5.58) (12.301,5.08) (12.618,4.63) (12.936,4.22) (13.253,3.84) (13.571,3.50) (13.888,3.19) (14.206,2.91) (14.523,2.65) (14.841,2.41) (15.158,2.20) (15.476,2.00) (15.793,1.82) (16.111,1.66) (16.429,1.51) (16.746,1.38) (17.064,1.25)};
\addlegendentry{NoMaD}
\addplot[mypurple, thick, smooth] coordinates {(1.505,100.00) (1.822,83.94) (2.140,70.04) (2.457,58.13) (2.775,48.04) (3.092,39.55) (3.410,32.46) (3.727,26.57) (4.045,21.70) (4.362,17.70) (4.680,14.41) (4.998,11.72) (5.315,9.52) (5.633,7.73) (5.950,6.27) (6.268,5.08) (6.585,4.12) (6.903,3.34) (7.220,2.70) (7.538,2.19) (7.855,1.77) (8.173,1.43) (8.490,1.16) (8.808,0.94) (9.125,0.76) (9.443,0.62) (9.760,0.50) (10.078,0.40) (10.395,0.33) (10.713,0.26) (11.031,0.21) (11.348,0.17) (11.666,0.14) (11.983,0.11) (12.301,0.09) (12.618,0.07) (12.936,0.06) (13.253,0.05) (13.571,0.04) (13.888,0.03) (14.206,0.03) (14.523,0.02) (14.841,0.02) (15.158,0.01) (15.476,0.01) (15.793,0.01) (16.111,0.01) (16.429,0.01) (16.746,0.00) (17.064,0.00)};
\addlegendentry{PlaceNav}
\addplot[myred, thick, smooth] coordinates {(1.505,100.00) (1.822,90.47) (2.140,81.60) (2.457,73.39) (2.775,65.84) (3.092,58.93) (3.410,52.63) (3.727,46.91) (4.045,41.73) (4.362,37.07) (4.680,32.88) (4.998,29.13) (5.315,25.77) (5.633,22.78) (5.950,20.12) (6.268,17.75) (6.585,15.65) (6.903,13.79) (7.220,12.14) (7.538,10.69) (7.855,9.40) (8.173,8.27) (8.490,7.27) (8.808,6.39) (9.125,5.61) (9.443,4.93) (9.760,4.33) (10.078,3.80) (10.395,3.34) (10.713,2.93) (11.031,2.57) (11.348,2.26) (11.666,1.98) (11.983,1.74) (12.301,1.52) (12.618,1.34) (12.936,1.17) (13.253,1.03) (13.571,0.90) (13.888,0.79) (14.206,0.69) (14.523,0.61) (14.841,0.53) (15.158,0.47) (15.476,0.41) (15.793,0.36) (16.111,0.32) (16.429,0.28) (16.746,0.24) (17.064,0.21)};
\addlegendentry{ViNT}
    \nextgroupplot[
      xlabel={Init. Distance (m)},
      ylabel={\small Relative Loc. Acc.},
      yticklabels={},  
      legend pos=south west,     
      xmin=1,
      xmax=11,
  ] 
\addplot[myorange, thick, smooth] coordinates {(1.513,100.00) (1.831,94.57) (2.148,89.41) (2.465,84.50) (2.783,79.84) (3.100,75.42) (3.417,71.22) (3.735,67.25) (4.052,63.47) (4.369,59.90) (4.687,56.52) (5.004,53.32) (5.322,50.29) (5.639,47.42) (5.956,44.71) (6.274,42.15) (6.591,39.73) (6.908,37.45) (7.226,35.29) (7.543,33.25) (7.860,31.32) (8.178,29.51) (8.495,27.79) (8.812,26.18) (9.130,24.65) (9.447,23.21) (9.764,21.86) (10.082,20.58) (10.399,19.37) (10.717,18.24) (11.034,17.17) (11.351,16.16) (11.669,15.21) (11.986,14.31) (12.303,13.47) (12.621,12.68) (12.938,11.93) (13.255,11.22) (13.573,10.56) (13.890,9.94) (14.207,9.35) (14.525,8.80) (14.842,8.27) (15.159,7.78) (15.477,7.32) (15.794,6.89) (16.112,6.48) (16.429,6.10) (16.746,5.73) (17.064,5.39)};
\addplot[myteal, thick, smooth] coordinates {(1.513,100.00) (1.831,89.57) (2.148,80.14) (2.465,71.64) (2.783,63.99) (3.100,57.11) (3.417,50.94) (3.735,45.40) (4.052,40.45) (4.369,36.01) (4.687,32.05) (5.004,28.52) (5.322,25.36) (5.639,22.55) (5.956,20.04) (6.274,17.81) (6.591,15.82) (6.908,14.06) (7.226,12.48) (7.543,11.08) (7.860,9.84) (8.178,8.74) (8.495,7.75) (8.812,6.88) (9.130,6.11) (9.447,5.42) (9.764,4.81) (10.082,4.27) (10.399,3.79) (10.717,3.36) (11.034,2.98) (11.351,2.64) (11.669,2.35) (11.986,2.08) (12.303,1.85) (12.621,1.64) (12.938,1.45) (13.255,1.29) (13.573,1.14) (13.890,1.01) (14.207,0.90) (14.525,0.80) (14.842,0.71) (15.159,0.63) (15.477,0.56) (15.794,0.49) (16.112,0.44) (16.429,0.39) (16.746,0.34) (17.064,0.31)};
\addplot[mypurple, thick, smooth] coordinates {(1.513,100.00) (1.831,83.99) (2.148,70.32) (2.465,58.72) (2.783,48.92) (3.100,40.67) (3.417,33.76) (3.735,27.99) (4.052,23.18) (4.369,19.17) (4.687,15.85) (5.004,13.09) (5.322,10.81) (5.639,8.92) (5.956,7.36) (6.274,6.07) (6.591,5.00) (6.908,4.13) (7.226,3.40) (7.543,2.80) (7.860,2.31) (8.178,1.90) (8.495,1.57) (8.812,1.29) (9.130,1.06) (9.447,0.88) (9.764,0.72) (10.082,0.59) (10.399,0.49) (10.717,0.40) (11.034,0.33) (11.351,0.27) (11.669,0.23) (11.986,0.19) (12.303,0.15) (12.621,0.13) (12.938,0.10) (13.255,0.09) (13.573,0.07) (13.890,0.06) (14.207,0.05) (14.525,0.04) (14.842,0.03) (15.159,0.03) (15.477,0.02) (15.794,0.02) (16.112,0.01) (16.429,0.01) (16.746,0.01) (17.064,0.01)};
\addplot[myred, thick, smooth] coordinates {(1.513,100.00) (1.831,91.99) (2.148,84.58) (2.465,77.74) (2.783,71.42) (3.100,65.59) (3.417,60.21) (3.735,55.26) (4.052,50.70) (4.369,46.50) (4.687,42.64) (5.004,39.09) (5.322,35.83) (5.639,32.83) (5.956,30.08) (6.274,27.56) (6.591,25.24) (6.908,23.12) (7.226,21.17) (7.543,19.38) (7.860,17.75) (8.178,16.24) (8.495,14.87) (8.812,13.61) (9.130,12.45) (9.447,11.40) (9.764,10.43) (10.082,9.54) (10.399,8.73) (10.717,7.99) (11.034,7.31) (11.351,6.69) (11.669,6.12) (11.986,5.60) (12.303,5.12) (12.621,4.68) (12.938,4.28) (13.255,3.92) (13.573,3.58) (13.890,3.28) (14.207,3.00) (14.525,2.74) (14.842,2.51) (15.159,2.29) (15.477,2.10) (15.794,1.92) (16.112,1.75) (16.429,1.60) (16.746,1.47) (17.064,1.34)};
\end{groupplot}%
\node[above=0.1cm] at ($(group c1r1.north)!0.5!(group c2r1.north)$) {\ref*{sharedlegend}};%
\end{tikzpicture}%
    \vspace{-0.8cm}
    \caption{Relative success rate (SR) for all methods on off-trajectory initialization over initialization distance (left), compared to subgoal localization accuracy for baseline methods (right). \methodname{} does not perform discrete subgoal selection and is therefore omitted from the right panel.}
    \label{fig:init_dist_per}
    \vspace{-0.4cm}
\end{figure}
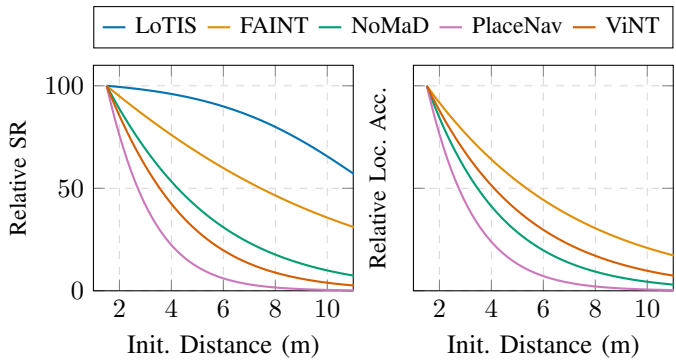
A common failure mode for visual navigation is the inability to recover when the robot begins far from the reference trajectory. As shown in the results, baseline performance drops sharply in the "Off-Trajectory" setting. For example, ViNT’s success rate on HM3D falls from 65.7\% to 24.0\%.

Our model remains robust in these settings, maintaining 88.2\% SR on Gibson. We observe lower off-trajectory performance on HM3D compared to Gibson (74.0\% vs.\ 88.2\%), which we attribute to HM3D's more cluttered environments. We note that the naive yaw controller at times causes the robot to get stuck behind small obstacles while attempting to return to the trajectory. By integrating basic obstacle avoidance, our success rate increases to 98.7\% (Gibson) and 95.1\% (HM3D).

This performance suggests that our model's predictions remain accurate even from distant viewpoints where baselines get lost. We further explore this in Fig.~\ref{fig:init_dist_per}, which shows the localization accuracy of the baselines (right), measured by correctly identifying the closest image of the reference trajectory, compared to their respective localization accuracy for on-trajectory following. We observe that this accuracy drops rapidly as the initialization distance increases, which leads to substantial degradation of performance in SR (left). By contrast, \methodname{}’s ability to localize the full trajectory appears to be substantially more robust, resulting in less sensitivity with respect to the initialization distance.
\subsubsection{Camera Mismatch Robustness}
The "Cross" columns in Table~\ref{tab:unified_results} evaluate each method's ability to handle mismatches in camera FOV, aspect ratio and camera mounting height. We observe a significant performance drop for all baselines here, e.g. ViNT's performance on Gibson drops from 70.4\% to 21.1\% in SR when the camera changes.

In contrast, \methodname{} maintains high performance across all evaluations (e.g. 83.6\% SR on Gibson, and 98.0\% when paired with obstacle avoidance). We attribute this to our cross-trajectory training strategy, which likely helps increase robustness for the model against camera mismatch between query and reference trajectory. We note that our method is more sensitive to large height mismatches than FOV or AR mismatch, as this can lead to the trajectory being outside the field of view of the robot which may result in the robot getting lost. We provide a more detailed analysis of each method's sensitivity to different parameter mismatches in Appendix~\ref{s_sec:sim_eval}.
\subsubsection{Backward Navigation}
\begin{figure*}[t]
\centering
\def\svgwidth{\textwidth}
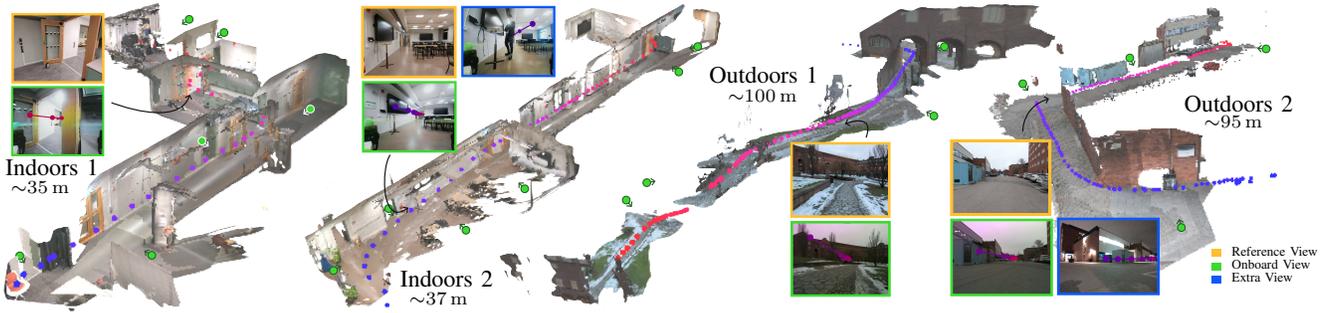
\vspace{-0.6cm}
\caption{Four trajectories used for real-world evaluation. 
Each reference trajectory starts at \protect\startpos~and ends at \protect\finalpos, 
with initial experiment positions shown as \protect\initpos. 
We present an offline-computed reconstruction of the environments~\cite{murai2024_mast3rslam} 
alongside representative views from: 
\protect\refview{reference trajectory camera}, 
\protect\onboardview{on-board camera} (analog FPV for indoors, RealSense~D455 for outdoors), 
and \protect\extraview{additional robustness study viewpoints}. 
Our model's trajectory predictions for the on-board cameras are overlaid in the corresponding views.}
\vspace{-0.4cm}
\label{fig:realworld_setup}
\end{figure*}
We evaluate the methods on the "To Start" task, where the robot must navigate to the first image of the reference trajectory. To successfully do this, the robot must be able to understand views that oppose the images of the reference trajectory. The results for this task are summarized in the "To Start" columns of Table~\ref{tab:unified_results}.

We observe a significant performance gap between \methodname{} and the baseline methods in this setting. On the Gibson dataset (On-Trajectory/Matched),\methodname{} achieves a 88.2\% success rate (SR) or 97.4\% when paired with obstacle avoidance. In contrast, the baselines largely fail: ViNT achieves 1.3\%, NoMaD 8.6\%, and FAINT 11.8\%. This trend remains consistent across the HM3D dataset, where \methodname{} maintains an 96.1\% SR while all baselines remain below 8\%. We note that here, even if subgoal selection is accurate, the baselines still largely fail due to the policy not being able to predict appropriate actions due to the robot's view opposing the chosen subgoal image.

When moving to the most challenging Off-Trajectory+Cross-Camera setting for backward navigation, \methodname{} performance experiences a moderate decrease but remains mostly successful, with success rates between 65.2\% and 86.8\% (86.8\% to 90.8\% when paired with obstacle avoidance). Meanwhile, the success rates of the baseline methods stay within the 2\% to 13\% range.

We note that \methodname{} achieves this without being trained on explicit backward traversal demonstrations. Backward traversal is difficult for subgoal-based policies because the current view and selected reference image may correspond to opposing travel directions, yielding limited visual overlap. Moreover, their action policies are trained on same-trajectory current-goal-action tuples and therefore do not observe such opposing-view pairs during training. In contrast, our cross-trajectory training does not require action annotations and naturally includes query views from other trajectories that observe the reference trajectory from different directions. Combined with joint trajectory processing, this allows \methodname{} to provide meaningful guidance even when individual frame-to-frame matches are ambiguous.

\subsubsection{Navigation to Arbitrary Goals (Any Point)}
The Any Point setting represents a more general usage of a reference trajectory where the robot is initialized off-trajectory and tasked to reach a certain point on the reference trajectory. This task serves as a summary for each method's performance as it requires strong capabilities in each of the aforementioned tasks. As shown in the rightmost columns of Table~\ref{tab:unified_results}, \methodname{} maintains high performance in this setting, achieving an 85.5\% SR on Gibson and 71.1\% on HM3D (increasing to $>$94\% with obstacle avoidance).

In summary, the results demonstrate that \methodname{} paired with only a simple yaw controller achieves substantially higher success rates than the baselines across all evaluations, while enabling backward traversal where baselines largely fail. When further paired with reactive collision avoidance, this leads to robust overall performance with over 95\% success rates across most experiments.
\section{Real-World Experiments}
We evaluate our method in real-world scenarios to answer:
\begin{itemize}
\item[\textbf{Q1:}] How does the performance transfer from simulation to real-world deployment?
\item[\textbf{Q2:}] How well does our method allow transfer to diverse embodiments (quadrotor, quadruped) from phone-recorded trajectories?
\item[\textbf{Q3:}] How do environmental variations, such as dynamic occlusions and day-night lighting changes, impact navigation performance?
\end{itemize}
\begin{figure}[h]
\centering
\def\svgwidth{\columnwidth}
\begingroup%
  \makeatletter%
  \providecommand\color[2][]{%
    \errmessage{(Inkscape) Color is used for the text in Inkscape, but the package 'color.sty' is not loaded}%
    \renewcommand\color[2][]{}%
  }%
  \providecommand\transparent[1]{%
    \errmessage{(Inkscape) Transparency is used (non-zero) for the text in Inkscape, but the package 'transparent.sty' is not loaded}%
    \renewcommand\transparent[1]{}%
  }%
  \providecommand\rotatebox[2]{#2}%
  \newcommand*\fsize{\dimexpr\f@size pt\relax}%
  \newcommand*\lineheight[1]{\fontsize{\fsize}{#1\fsize}\selectfont}%
  \ifx\svgwidth\undefined%
    \setlength{\unitlength}{599.59814838bp}%
    \ifx\svgscale\undefined%
      \relax%
    \else%
      \setlength{\unitlength}{\unitlength * \real{\svgscale}}%
    \fi%
  \else%
    \setlength{\unitlength}{\svgwidth}%
  \fi%
  \global\let\svgwidth\undefined%
  \global\let\svgscale\undefined%
  \makeatother%
  \begin{picture}(1,0.64025548)%
    \lineheight{1}%
    \setlength\tabcolsep{0pt}%
    \put(0,0){\includegraphics[width=\unitlength,page=1]{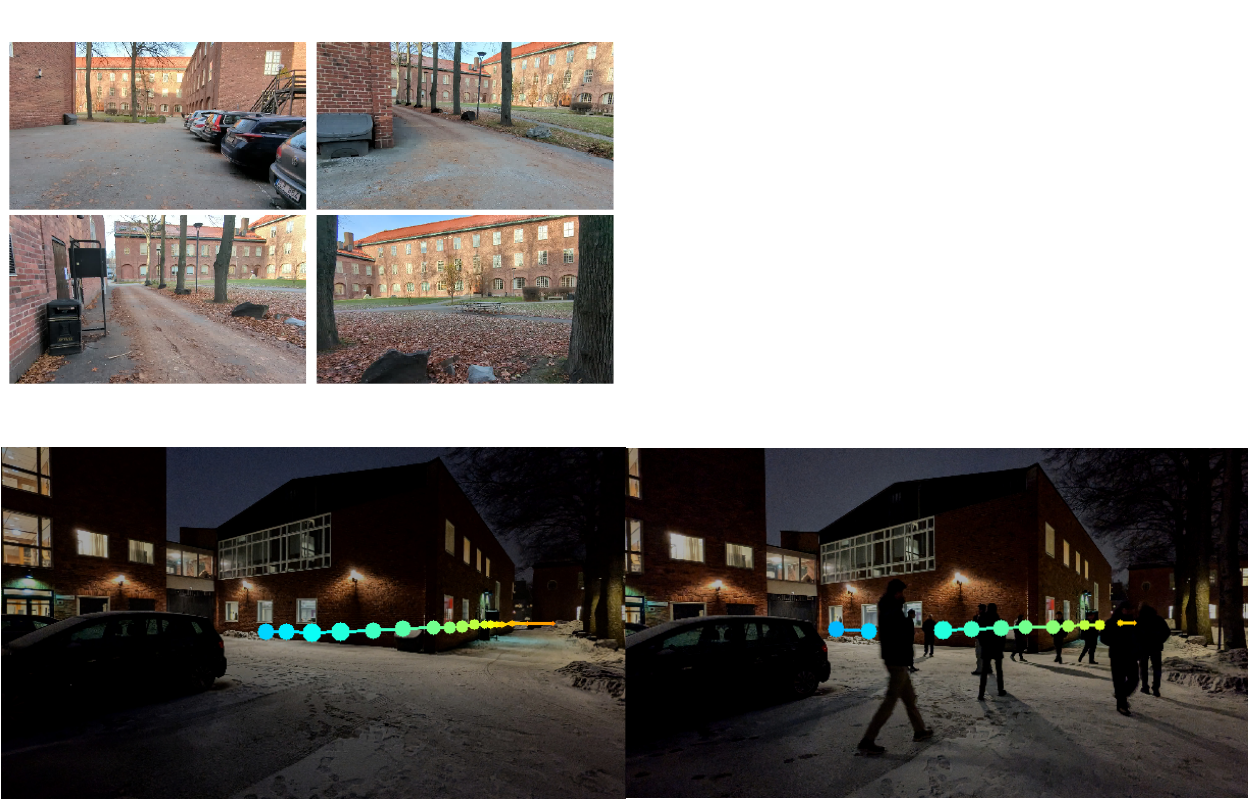}}%
    \put(0.03231386,0.62885135){\color[rgb]{0,0,0}\makebox(0,0)[lt]{\lineheight{1.25}\smash{\begin{tabular}[t]{l}\small Reference Trajectory $\mathcal{T}$\end{tabular}}}}%
    \put(0.60740949,0.62885135){\color[rgb]{0,0,0}\makebox(0,0)[lt]{\lineheight{1.25}\smash{\begin{tabular}[t]{l}\small Seasonal Change\end{tabular}}}}%
    \put(0.09712875,0.29879494){\color[rgb]{0,0,0}\makebox(0,0)[lt]{\lineheight{1.25}\smash{\begin{tabular}[t]{l}\small + Lighting Change\end{tabular}}}}%
    \put(0.60740949,0.29793009){\color[rgb]{0,0,0}\makebox(0,0)[lt]{\lineheight{1.25}\smash{\begin{tabular}[t]{l}\small + Occlusions\end{tabular}}}}%
    \put(0,0){\includegraphics[width=\unitlength,page=2]{generalization.pdf}}%
  \end{picture}%
\endgroup%

\vspace{-0.6cm}
\caption{Impact of environment changes on the predictions of our model with respect to a reference trajectory recorded on a sunny autumn day. Top Right: Seasonal Change, Bottom Left: Seasonal and day-night change, Bottom Right: Seasonal, day-night change and people occluding the view.}
\vspace{-0.4cm}
\label{fig:generalization}
\end{figure}
\subsection{Evaluation Setup}
We collect four reference trajectories using a handheld Google Pixel 6 smartphone: two indoors and two outdoors, see Fig.~\ref{fig:realworld_setup}. We evaluate on a Crazyflie quadrotor with an analog FPV camera with MPPI control (Sec.~\ref{sec:control}) for indoors trajectories, and on a Boston Dynamics Spot equipped with a Realsense~D455 with our yaw controller (Sec.~\ref{sec:control}), but leave its on-board collision avoidance enabled.
All trials initialize off-trajectory. We compare against FAINT~\cite{suomela2025faint}, the strongest baseline in the off-trajectory cross-camera simulation setting. Each trajectory is evaluated with 6 trials (indoors) or 3 trials (outdoors) per direction, totaling 36 runs per method.

For the Spot, we run our method on a Jetson Orin AGX, and for the Crazyflie, we run computation off-board on a desktop with an RTX 5090. We provide videos of all real-world evaluations on the project page \projectpage.
\subsection{Runtime}
\methodname{} uses an asymmetric deployment pipeline. When a reference trajectory is provided, all reference images are processed once by the frozen DINOv3 backbone and trajectory encoder $\mathcal{E}_T$. The resulting trajectory features $F_T$ are kept in memory and reused throughout navigation. This offline step is performed before online inference begins. During online deployment, each incoming camera image is processed only by the query encoder, query-trajectory fusion module, and prediction head, producing image-space trajectory predictions for the controller.

We report timings for both real-world compute platforms. On the Jetson Orin AGX used onboard Spot, offline trajectory encoding takes $120\,\mathrm{ms}$ for a trajectory of up to 40 frames, while online query encoding and decoding takes $45\,\mathrm{ms}$ per frame, yielding approximately $22\,\mathrm{Hz}$ control-rate inference. On the RTX 5090 desktop used for the Crazyflie experiments, offline trajectory encoding takes $15\,\mathrm{ms}$ and online inference takes $6\,\mathrm{ms}$ per frame, yielding approximately $160\,\mathrm{Hz}$. The offline timing includes DINOv3 reference feature extraction and the trajectory encoder forward pass, but is not part of the per-frame control loop. All reported timings use trajectories subsampled to 40 frames.
\subsection{Main Results}
\begin{table}[h]
\centering
\caption{Real-world navigation results. All trials use phone-recorded trajectories with off-trajectory initialization. Indoors: Crazyflie with MPPI. Outdoors: Spot with yaw controller.}
\label{tab:rw_results}
\vspace{-0.2cm}
\small
\setlength{\tabcolsep}{3.5pt}
\begin{tabular}{@{}lcccccccc@{}}
\toprule
 & \multicolumn{2}{c}{Indoors 1} & \multicolumn{2}{c}{Indoors 2} & \multicolumn{2}{c}{Outdoors 1} & \multicolumn{2}{c}{Outdoors 2} \\
\cmidrule(lr){2-3} \cmidrule(lr){4-5} \cmidrule(lr){6-7} \cmidrule(l){8-9}
 & Fwd & Bwd & Fwd & Bwd & Fwd & Bwd & Fwd & Bwd \\
\midrule
FAINT~\cite{suomela2025faint}   & 3/6 & 0/6 & 1/6 & 1/6 & 1/3 & 0/3 & \textbf{3/3} & 0/3 \\
\methodname{} (Ours) & \textbf{6/6} & \textbf{5/6} & \textbf{6/6} & \textbf{6/6} & \textbf{3/3} & \textbf{3/3} & \textbf{3/3} & \textbf{3/3} \\
\bottomrule
\end{tabular}
\vspace{-0.3cm}
\end{table}
As shown in Table~\ref{tab:rw_results}, our method achieves 100\% success on forward navigation across all environments, while FAINT succeeds in only 27.8\% of trials. This matches, or outperforms the results obtained in simulation. For indoors, we attribute the improved real-world performance compared to sim to the MPPI's ability to keep the predicted trajectory centered in view by actively matching the reference height.
We note that for fairness, we manually adjust the drone's flight height to match the recording height for FAINT since FAINT solely provides actions in the horizontal plane.
We show that the yaw controller is sufficient for the evaluated outdoors settings due to larger open spaces where Spot's internal reactive collision avoidance is sufficient, though it was not required to intervene during our evaluations. For both indoors and outdoors, we observe that FAINT being unable to reliably localize from off-trajectory initializations, is a main cause of failure.

The gap widens further on backward traversal: our method maintains 94.4\% success (17/18), whereas FAINT largely fails (5.6\%). These results match our simulation findings and suggest that our cross-trajectory training strategy and joint processing of the full trajectory enables the challenging task of backward traversal.
Though performance remains consistent overall, we occasionally encounter views during backward traversal where our model can no longer match the view, and thus predicts no visible points. For the indoors setting, the MPPI is warm-started by the previous solution and will thus continue following the previous solutions, usually leading to better views and recovery within a few steps. For outdoors, we hypothesize that larger open spaces are less prone to produce views with no or little overlap with the trajectory.

\begin{table}[h]
\centering
\vspace{-0.2cm}
\caption{Robustness to environmental changes. \textbf{Crowded Env}: Indoors 2 with and without people occluding the view. \textbf{Day$\rightarrow$Night}: Outdoors 2 evaluated at night with scene changes.}
\label{tab:rw_ablation}
\small
\setlength{\tabcolsep}{6pt}
\begin{tabular}{@{}lcccc@{}}
\toprule
 & \multicolumn{2}{c}{\textbf{Crowded Env}} & \multicolumn{2}{c}{\textbf{Day}$\rightarrow$\textbf{Night}} \\
\cmidrule(lr){2-3} \cmidrule(l){4-5}
 & Clean & +People & Day & Night \\
\midrule
Forward  & 6/6 & 5/6 & 3/3 & 3/3 \\
Backward & 6/6 & 5/6 & 3/3 & 2/3 \\
\bottomrule
\end{tabular}
\vspace{-0.3cm}
\end{table}
\subsection{Robustness Experiments}
Table~\ref{tab:rw_ablation} evaluates robustness under challenging conditions, and we provide an example of our model queried for representative challenging conditions in Fig.~\ref{fig:generalization}. We study two changes for the navigating agent: 1.) We repeat the Indoors 2 evaluation with people present in the environment, at times occluding the robot's view and blocking its path, and 2.) We repeat the Outdoors 2 evaluation but with changes in scene (crowded parking lot is now empty) and at night time. Note that the reference trajectory remains the same as in our original evaluation, i.e. no people for indoors and a crowded parking lot at daytime for outdoors.
When people walk through the scene and temporarily occlude the camera view, our method maintains high success (5/6 forward, 5/6 backward). The model's predictions remain accurate, even if a large proportion of the view is occluded and correctly predicts points that are still visible. The one additional failure case can be attributed to people walking in the drone's path in a way that the drone needs to steer away from the reference trajectory, ultimately losing view of the trajectory and being unable to recover (see videos on the project page). For day-to-night transfer with simultaneous scene changes (trajectory recorded in a daytime parking lot with cars, evaluated at night with the lot empty), we achieve 6/6 forward and 5/6 backward. The one failure case for day-night can be attributed to one initial condition whose view was largely impacted by the scene changes, which leads to the system not being able to recognize any points of the trajectory. While we observe some degradation in the quality of the predictions for the day-night transfer, the results demonstrate that our learned representation is robust to appearance and geometric variation.
\vspace{-0.2cm}
\section{Conclusion}
We presented \methodname{}, a model for visual navigation that predicts where the reference trajectory would appear in the robot's current view. Our approach provides zero-shot guidance for different embodiments, and enables backward traversal and robustness to camera mismatch---capabilities difficult for prior end-to-end methods. Experiments demonstrate 94-98\% success on forward navigation across diverse embodiments both in simulation and real-world, and $5\times$ improvement on backward traversal, with real-world deployment confirming transfer to both quadrotor and quadruped platforms using phone-recorded trajectories. For the first time, we used a single RGB reference trajectory in the general setting where the robot can navigate from anywhere in the vicinity to any point on the trajectory, both forward and backward. 

\textbf{Limitations.} Our method requires visual overlap between the current view and some portion of the reference trajectory; failures occur when this overlap is lost, particularly during backward traversal around sharp corners or under extreme viewpoint differences. Performance degrades with large camera height mismatches (see Appendix~\ref{s_sec:sim_eval}). Our real-world evaluation is limited to four environments, and does not exhaustively cover the diversity of challenging scenarios (e.g., unstructured environments, or highly repetitive environments) where failure modes may differ. We evaluated trajectories up to ${\sim}100\, \mathrm{m}$ subsampled to 40 frames. Longer trajectories require chunking to ensure coverage, which we demonstrate at kilometer-scale on the project page but do not systematically evaluate. \methodname{} assumes a pre-recorded trajectory with visual overlap to the current view, and navigation to entirely unseen goals remains outside our setting.

\textbf{Future work.} Incorporating temporal memory could help recover when the trajectory temporarily leaves the field of view. Learning to provide guidance even without a clear view of the trajectory could address the limitation of height mismatch. Finally, combining image-space guidance with semantic understanding could enable navigation to functional goals (e.g., ``the kitchen'') rather than trajectory indices.

\bibliographystyle{plainnat}
\bibliography{references}
\newpage
\clearpage
\setcounter{section}{0}
\setcounter{subsection}{0}
\setcounter{figure}{0}
\setcounter{table}{0}

\renewcommand{\thesection}{\Alph{section}}
\renewcommand{\thesubsection}{\Alph{section}.\arabic{subsection}}
\renewcommand{\thefigure}{A\arabic{figure}}
\renewcommand{\thetable}{A\arabic{table}}

\renewcommand{\theHsection}{appendix.\Alph{section}}
\renewcommand{\theHsubsection}{appendix.\Alph{section}.\arabic{subsection}}
\renewcommand{\theHfigure}{appendix.A\arabic{figure}}
\renewcommand{\theHtable}{appendix.A\arabic{table}}
\twocolumn[
\begin{center}
{\LARGE \bfseries Appendix\\[0.5em]}
\vspace{1em}
\end{center}
]
\section{Overview}
This appendix provides additional implementation details, extended results, and video demonstrations for our paper.
The appendix is organized as follows:
\begin{itemize}
    \item video demonstrations (project page), Sec.~\ref{s_sec:videos}
    \item an ablation on full-trajectory vs. frame-wise processing, Sec.~\ref{s_sec:ablation_ft}
    \item the parameter statistics of simulation evaluation dataset, Sec.~\ref{s_sec:stats}
    \item details on performance under camera mismatch for each varied camera parameter, Sec.~\ref{s_sec:sim_eval}
    \item model architecture and training details, Sec.~\ref{s_sec:impl_details}
    \item qualitative comparison with MASt3R and VGGT for navigation, Sec.~\ref{s_sec:vggt_study}
    \item navigation implementation details, Sec.~\ref{s_sec:contr_details}
\end{itemize}
\section{Video Demonstrations}
\label{s_sec:videos}
We highly encourage the reader to review the video demonstrations available on the project page: \projectpage

The videos include:
\begin{itemize}
    \item \textbf{Real-World Navigation}: All 36 trials (forward and backward) on quadrotor on two indoor, and quadruped on two outdoors environments for \methodname{} (Table~\ref{tab:rw_results})
    \item \textbf{Robustness Studies}: Day-to-night transfer, dynamic occlusions, and environmental changes (Table~\ref{tab:rw_ablation})
    \item \textbf{Kilometer-Scale Demonstration}: Long-range inference with trajectory chunking
\end{itemize}
\section{Ablation: Full-Trajectory vs. Frame-wise Processing}
\label{s_sec:ablation_ft}
To investigate the importance of joint processing the full reference trajectory rather than matching the query view with each frame individually, we conduct an ablation comparing our default model (\textbf{Full}) against a frame-wise baseline (\textbf{Single}).

\textbf{Configuration.} In the \textbf{Single} configuration, we modify the Trajectory Encoder $\mathcal{E}_T$ to process each reference frame $I_i \in \mathcal{T}$ independently. We achieve this by restricting the attention mechanism in $\mathcal{E}_T$: tokens are only allowed to attend to other tokens within the same frame, effectively treating a trajectory of length $N$ as $N$ independent trajectories of length 1. 

The Query-Trajectory Fusion module $\mathcal{F}_{q\mathcal{T}}$ remains unchanged. However, we note that by design, this module performs cross-attention where trajectory tokens attend to the query image tokens. Consequently, there is no communication between different reference trajectory frames within the fusion layers. 
The only stage where information is exchanged across the trajectory sequence in the \textbf{Single} model is within the \textit{Prediction Head}, via self-attention over the summary tokens $c_i$. We retain this final mixing stage because the model is required to output normalized distances $d_i$ (where $d=1$ corresponds to the farthest visible point), an operation that inherently requires access to the distribution of predictions across the sequence.

In the "Single" model, the patch-level features of frame $I_i$ cannot inform the representation of frame $I_j$ during encoding or fusion; temporal consistency can only be recovered by the prediction head using the compressed summary tokens.
\begin{table}[h]
\centering
\setlength{\tabcolsep}{1.8pt}
\renewcommand{\arraystretch}{1.0}
\caption{Ablation on processing the full trajectory jointly (\methodname{}-F), or processing the trajectory as individual frames before matching (\methodname{}-S). We report Success Rate (SR). The \textit{Diff} row shows the drop in SR when using Single vs Full.}
\vspace{-0.3cm}
\label{tab:ablation}
\begin{tabular}{c l cc cc | cc cc }
\toprule
 & & \multicolumn{4}{c|}{\textbf{To End (Forward)}} & \multicolumn{4}{c}{\textbf{To Start (Backward)}}  \\
 & & \multicolumn{2}{c}{On-Trajectory} & \multicolumn{2}{c}{Off-Trajectory} & \multicolumn{2}{c}{On-Trajectory} & \multicolumn{2}{c}{Off-Trajectory} \\
\cmidrule(lr){3-4} \cmidrule(lr){5-6} \cmidrule(lr){7-8} \cmidrule(lr){9-10} 
 & \textbf{Method} & Matched & Cross & Matched & Cross & Matched & Cross & Matched & Cross  \\
\midrule
\multirow{3}{*}{\rotatebox[origin=c]{90}{\textbf{Gibson}}} 
  & \methodname{}-F   & \textbf{100.0} & \textbf{98.0} & \textbf{98.7} & \textbf{94.7} & \textbf{97.4} & \textbf{89.5} & \textbf{96.7} & \textbf{90.8} \\
  & \methodname{}-S & 95.4 & 91.4 & 86.8 & 79.6 & 69.7 & 67.8 & 71.1 & 57.9 \\ 
  & \textit{Diff}     & -4.6 & -6.6 & -11.9 & -15.1 & -27.7 & -21.7 & -25.6 & -32.9 \\
\midrule

\multirow{3}{*}{\rotatebox[origin=c]{90}{\textbf{HM3D}}}
  & \methodname{}-F & \textbf{100.0} & \textbf{97.5} & \textbf{95.1} & \textbf{92.2} & \textbf{96.1} & \textbf{84.8} & \textbf{92.6} & \textbf{86.8} \\
  & \methodname{}-S & 94.1 & 78.9 & 68.6 & 67.6 & 63.2 & 59.8 & 59.3 & 56.4 \\
  & \textit{Diff}     & -5.9 & -18.6 & -26.5 & -24.6 & -32.9 & -25.0 & -33.3 & -30.4 \\

\bottomrule
\end{tabular}
\vspace{-0.4cm}
\end{table}

\begin{figure}[h]
    \centering
    \definecolor{myblue}{HTML}{0173B2}
\definecolor{myorange}{HTML}{DE8F05}
\definecolor{myteal}{HTML}{029E73}
\definecolor{mypurple}{HTML}{CC78BC}
\definecolor{myred}{HTML}{D55E00}
\begin{tikzpicture}
\begin{axis}[
    xlabel={Initialization Distance from Path (m)},
    ylabel={Relative SR},
    grid=major,
    grid style={dashed, gray!30},
    legend pos=south west,
    legend cell align=left,
    ymin=0, ymax=110,
    xmin=1,
    xmax=11,
    width=1.0\columnwidth,
    height=5.5cm,
    legend columns=3,
    legend style={                                                                                                                                                                                                                                                 
      font=\small,                                                                                                                                                                                                                                               
      draw=black,  
      column sep=0.05cm,  
      legend image post style={scale=0.5},  
  },  
]
\addplot[myblue, dashed, thick, smooth] coordinates {(1.505,100.00) (1.822,99.65) (2.140,99.26) (2.457,98.82) (2.775,98.35) (3.092,97.82) (3.410,97.25) (3.727,96.61) (4.045,95.91) (4.362,95.15) (4.680,94.31) (4.998,93.39) (5.315,92.39) (5.633,91.30) (5.950,90.12) (6.268,88.84) (6.585,87.45) (6.903,85.96) (7.220,84.36) (7.538,82.64) (7.855,80.81) (8.173,78.86) (8.490,76.80) (8.808,74.63) (9.125,72.35) (9.443,69.98) (9.760,67.50) (10.078,64.95) (10.395,62.32) (10.713,59.64) (11.031,56.91) (11.348,54.14) (11.666,51.37) (11.983,48.59) (12.301,45.83) (12.618,43.11) (12.936,40.43) (13.253,37.82) (13.571,35.28) (13.888,32.82) (14.206,30.46) (14.523,28.20) (14.841,26.05) (15.158,24.01) (15.476,22.08) (15.793,20.27) (16.111,18.57) (16.429,16.99) (16.746,15.52) (17.064,14.15)};
\addlegendentry{\methodname{} (Full)}
\addplot[myblue, thick, smooth] coordinates {(1.505,100.00) (1.822,98.40) (2.140,96.76) (2.457,95.06) (2.775,93.32) (3.092,91.53) (3.410,89.70) (3.727,87.82) (4.045,85.90) (4.362,83.95) (4.680,81.97) (4.998,79.95) (5.315,77.91) (5.633,75.84) (5.950,73.76) (6.268,71.66) (6.585,69.54) (6.903,67.43) (7.220,65.30) (7.538,63.19) (7.855,61.07) (8.173,58.97) (8.490,56.88) (8.808,54.82) (9.125,52.77) (9.443,50.75) (9.760,48.76) (10.078,46.81) (10.395,44.89) (10.713,43.01) (11.031,41.17) (11.348,39.37) (11.666,37.62) (11.983,35.92) (12.301,34.27) (12.618,32.67) (12.936,31.12) (13.253,29.62) (13.571,28.17) (13.888,26.78) (14.206,25.44) (14.523,24.15) (14.841,22.91) (15.158,21.72) (15.476,20.58) (15.793,19.49) (16.111,18.45) (16.429,17.46) (16.746,16.51) (17.064,15.61)};
\addlegendentry{\methodname{} (Single)}
\addplot[myorange, dashed, thick, smooth] coordinates {(1.505,100.00) (1.822,96.73) (2.140,93.52) (2.457,90.38) (2.775,87.29) (3.092,84.28) (3.410,81.33) (3.727,78.45) (4.045,75.64) (4.362,72.90) (4.680,70.23) (4.998,67.63) (5.315,65.10) (5.633,62.64) (5.950,60.25) (6.268,57.93) (6.585,55.69) (6.903,53.51) (7.220,51.40) (7.538,49.36) (7.855,47.38) (8.173,45.47) (8.490,43.62) (8.808,41.84) (9.125,40.12) (9.443,38.46) (9.760,36.86) (10.078,35.32) (10.395,33.84) (10.713,32.41) (11.031,31.03) (11.348,29.71) (11.666,28.43) (11.983,27.21) (12.301,26.03) (12.618,24.90) (12.936,23.82) (13.253,22.78) (13.571,21.78) (13.888,20.82) (14.206,19.90) (14.523,19.02) (14.841,18.18) (15.158,17.37) (15.476,16.60) (15.793,15.85) (16.111,15.14) (16.429,14.46) (16.746,13.81) (17.064,13.19)};
\addlegendentry{FAINT}
\addplot[myteal, dashed, thick, smooth] coordinates {(1.505,100.00) (1.822,92.68) (2.140,85.78) (2.457,79.30) (2.775,73.22) (3.092,67.54) (3.410,62.23) (3.727,57.29) (4.045,52.69) (4.362,48.42) (4.680,44.47) (4.998,40.81) (5.315,37.42) (5.633,34.30) (5.950,31.42) (6.268,28.76) (6.585,26.32) (6.903,24.08) (7.220,22.02) (7.538,20.12) (7.855,18.39) (8.173,16.80) (8.490,15.34) (8.808,14.00) (9.125,12.78) (9.443,11.66) (9.760,10.64) (10.078,9.71) (10.395,8.85) (10.713,8.07) (11.031,7.36) (11.348,6.71) (11.666,6.12) (11.983,5.58) (12.301,5.08) (12.618,4.63) (12.936,4.22) (13.253,3.84) (13.571,3.50) (13.888,3.19) (14.206,2.91) (14.523,2.65) (14.841,2.41) (15.158,2.20) (15.476,2.00) (15.793,1.82) (16.111,1.66) (16.429,1.51) (16.746,1.38) (17.064,1.25)};
\addlegendentry{NoMaD}
\addplot[mypurple, dashed, thick, smooth] coordinates {(1.505,100.00) (1.822,83.94) (2.140,70.04) (2.457,58.13) (2.775,48.04) (3.092,39.55) (3.410,32.46) (3.727,26.57) (4.045,21.70) (4.362,17.70) (4.680,14.41) (4.998,11.72) (5.315,9.52) (5.633,7.73) (5.950,6.27) (6.268,5.08) (6.585,4.12) (6.903,3.34) (7.220,2.70) (7.538,2.19) (7.855,1.77) (8.173,1.43) (8.490,1.16) (8.808,0.94) (9.125,0.76) (9.443,0.62) (9.760,0.50) (10.078,0.40) (10.395,0.33) (10.713,0.26) (11.031,0.21) (11.348,0.17) (11.666,0.14) (11.983,0.11) (12.301,0.09) (12.618,0.07) (12.936,0.06) (13.253,0.05) (13.571,0.04) (13.888,0.03) (14.206,0.03) (14.523,0.02) (14.841,0.02) (15.158,0.01) (15.476,0.01) (15.793,0.01) (16.111,0.01) (16.429,0.01) (16.746,0.00) (17.064,0.00)};
\addlegendentry{PlaceNav}
\addplot[myred, dashed, thick, smooth] coordinates {(1.505,100.00) (1.822,90.47) (2.140,81.60) (2.457,73.39) (2.775,65.84) (3.092,58.93) (3.410,52.63) (3.727,46.91) (4.045,41.73) (4.362,37.07) (4.680,32.88) (4.998,29.13) (5.315,25.77) (5.633,22.78) (5.950,20.12) (6.268,17.75) (6.585,15.65) (6.903,13.79) (7.220,12.14) (7.538,10.69) (7.855,9.40) (8.173,8.27) (8.490,7.27) (8.808,6.39) (9.125,5.61) (9.443,4.93) (9.760,4.33) (10.078,3.80) (10.395,3.34) (10.713,2.93) (11.031,2.57) (11.348,2.26) (11.666,1.98) (11.983,1.74) (12.301,1.52) (12.618,1.34) (12.936,1.17) (13.253,1.03) (13.571,0.90) (13.888,0.79) (14.206,0.69) (14.523,0.61) (14.841,0.53) (15.158,0.47) (15.476,0.41) (15.793,0.36) (16.111,0.32) (16.429,0.28) (16.746,0.24) (17.064,0.21)};
\addlegendentry{ViNT}
\end{axis}%
\end{tikzpicture}%
    \caption{\methodname{} with full joint trajectory processing (Full) compared against frame-wise trajectory processing (Single), and all baselines. Processing the full-trajectory leads to better result when initialized farther, i.e. in more challenging views.}
    \label{fig:abl_full_frame}
    \vspace{-0.3cm}
\end{figure}
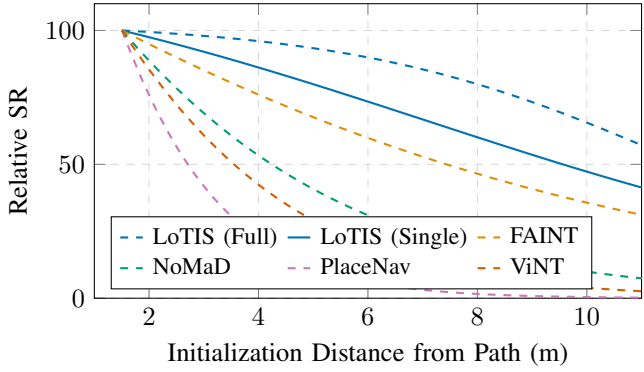

\textbf{Discussion.} The results are reported in Table~\ref{tab:ablation}. We make the following observations:
\begin{enumerate}
    \item \textbf{Degradation with Distance:} As shown in Fig.~\ref{fig:abl_full_frame}, while both methods perform well when initialized on the trajectory, the performance of "Single" degrades rapidly as the initialization distance increases. The degradation curve of "Single" nears those of the best baseline (FAINT), which also relies on pairwise matching or retrieval. By contrast, "Full" maintains significantly higher relative success rates at large distances ($>8$\,m). This suggests that early joint processing allows the trajectory encoder to learn a consistent geometric structure of the trajectory, providing robustness under more challenging viewpoints.
    \item \textbf{Backward Traversal:} In Table~\ref{tab:ablation}, the performance gap is most pronounced in the "Backward" setting ($\sim$30\% drop in SR). Backward traversal often results in views with very challenging viewpoints. We hypothesize that the "Single" model struggles here because the Query-Trajectory Fusion needs to match the query view pairwise with every view of the trajectory.
\end{enumerate}
These results support that joint trajectory-level processing, rather than independent pairwise frame matching, is particularly important for off-trajectory and backward traversal settings.
\section{Evaluation Dataset Statistics}
\label{s_sec:stats}
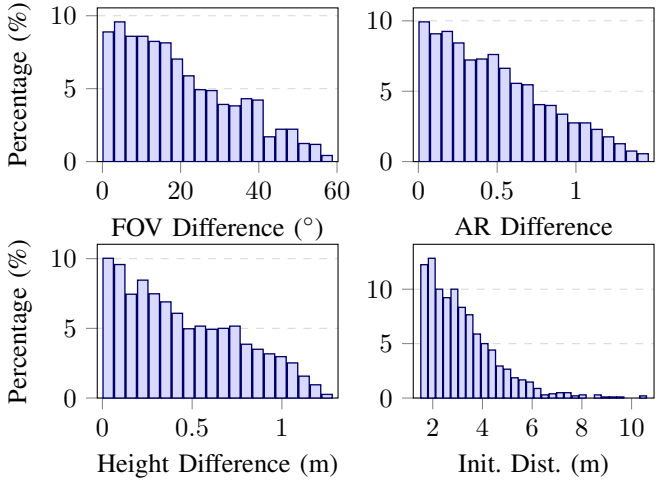
\begin{figure}[ht]
    \centering
    \begin{tikzpicture}
    \begin{groupplot}[
        group style={
            group size=2 by 2,
            horizontal sep=1.cm,
            vertical sep=1.1cm,
        },
        width=0.54\columnwidth,
        height=0.2\textwidth,
        ybar,
        ymin=0,
        ymajorgrids=true,
        grid style={dashed,gray!30},
        tick align=outside,
        tick pos=left,
        enlarge x limits=0.05,
        nodes near coords={}, 
        every node near coord/.append style={font=\tiny},
    ]

    \nextgroupplot[
        ylabel={Percentage (\%)},
        xlabel={FOV Difference ($^{\circ}$)},
        bar width=2.6520
    ]
    \addplot[
        fill=blue!15, 
        draw=blue!40!black, 
        line width=0.5pt
    ] coordinates {
        (1.4765, 8.89)
        (4.4232, 9.58)
        (7.3700, 8.59)
        (10.3167, 8.59)
        (13.2634, 8.24)
        (16.2101, 8.14)
        (19.1568, 7.03)
        (22.1036, 5.88)
        (25.0503, 4.93)
        (27.9970, 4.87)
        (30.9437, 3.92)
        (33.8904, 3.82)
        (36.8372, 4.31)
        (39.7839, 4.22)
        (42.7306, 1.70)
        (45.6773, 2.22)
        (48.6240, 2.22)
        (51.5708, 1.24)
        (54.5175, 1.18)
        (57.4642, 0.42)
    };

    \nextgroupplot[
        xlabel={AR Difference},
        bar width=0.0658
    ]
    \addplot[
        fill=blue!15, 
        draw=blue!40!black, 
        line width=0.5pt
    ] coordinates {
        (0.0369, 9.93)
        (0.1100, 9.08)
        (0.1831, 9.25)
        (0.2562, 8.43)
        (0.3294, 7.22)
        (0.4025, 7.29)
        (0.4756, 7.61)
        (0.5487, 6.63)
        (0.6219, 5.56)
        (0.6950, 5.46)
        (0.7681, 4.05)
        (0.8412, 3.99)
        (0.9144, 3.37)
        (0.9875, 2.75)
        (1.0606, 2.75)
        (1.1337, 2.29)
        (1.2069, 1.76)
        (1.2800, 1.27)
        (1.3531, 0.75)
        (1.4262, 0.56)
    };

    \nextgroupplot[
        ylabel={Percentage (\%)},
        xlabel={Height Difference (m)},
        bar width=0.0578
    ]
    \addplot[
        fill=blue!15, 
        draw=blue!40!black, 
        line width=0.5pt
    ] coordinates {
        (0.0324, 10.03)
        (0.0966, 9.58)
        (0.1607, 7.45)
        (0.2249, 8.46)
        (0.2891, 7.48)
        (0.3533, 6.90)
        (0.4175, 6.08)
        (0.4817, 4.97)
        (0.5458, 5.16)
        (0.6100, 4.93)
        (0.6742, 5.00)
        (0.7384, 5.16)
        (0.8026, 3.86)
        (0.8667, 3.50)
        (0.9309, 3.17)
        (0.9951, 2.97)
        (1.0593, 2.52)
        (1.1235, 1.57)
        (1.1876, 0.95)
        (1.2518, 0.26)
    };

    \nextgroupplot[
        xlabel={Init. Dist. (m)},
        bar width=0.2733
    ]
    \addplot[
        fill=blue!15, 
        draw=blue!40!black, 
        line width=0.5pt
    ] coordinates {
        (1.6565, 12.25)
        (1.9602, 12.84)
        (2.2638, 10.00)
        (2.5675, 9.22)
        (2.8711, 10.00)
        (3.1747, 8.33)
        (3.4784, 7.65)
        (3.7820, 5.88)
        (4.0857, 5.00)
        (4.3893, 4.41)
        (4.6930, 2.94)
        (4.9966, 2.65)
        (5.3003, 1.86)
        (5.6039, 1.67)
        (5.9076, 1.47)
        (6.2112, 0.88)
        (6.5149, 0.29)
        (6.8185, 0.39)
        (7.1222, 0.49)
        (7.4258, 0.49)
        (7.7294, 0.20)
        (8.0331, 0.29)
        (8.6404, 0.29)
        (8.9440, 0.10)
        (9.2477, 0.10)
        (9.5513, 0.10)
        (10.4623, 0.20)
    };

    \end{groupplot}%
\end{tikzpicture}%
    \caption{Dataset statistics across all evaluation scenes. Camera parameter differences (a-c) are computed for all query types, while off-trajectory distances (d) are computed only for the off-trajectory queries.}
    \label{fig:dataset_stats}
\end{figure}
In this section, we provide more details on the variations introduced in our simulation evaluation to test robustness. Fig.~\ref{fig:dataset_stats} illustrates the distribution of camera parameter mismatches and initialization offsets across the evaluation.

\section{Per-Parameter Camera Mismatch Analysis}
\label{s_sec:sim_eval}
In Sec.~\ref{sec:sim_experiments} of the main paper, we evaluate the impact of camera parameter mismatches on each method's performance. Here, we provide further details on the impact of each specific parameter.
\begin{figure}[h]
    \centering
    \input{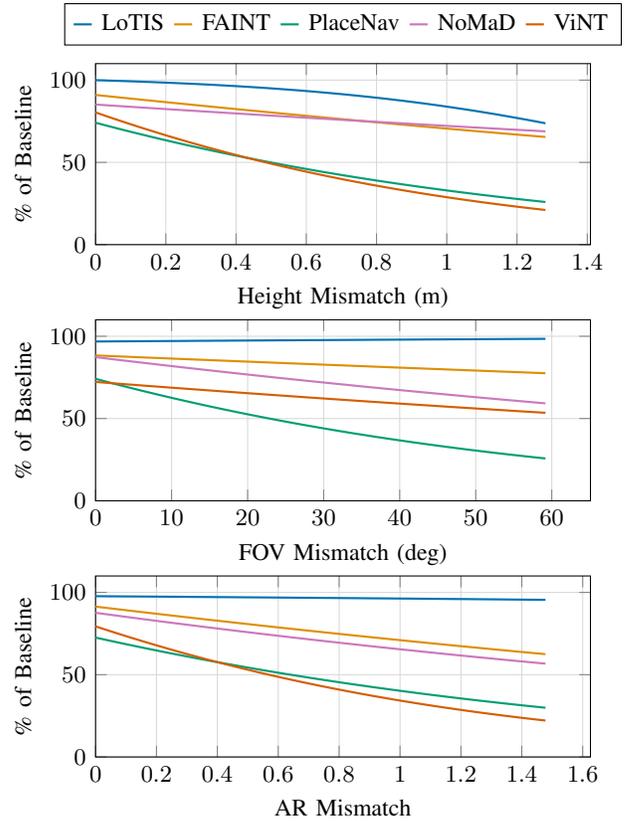}
    \caption{Performance degradation vs. camera parameter mismatch.}
    \label{fig:param-mismatch}
    \vspace{-0.5cm}
\end{figure}
Fig.~\ref{fig:param-mismatch} shows the performance of each method over the magnitude of each parameter's mismatch, compared to the performance achieved with no camera mismatch.

\methodname{} is largely insensitive to mismatches in FOV and aspect ratio (AR), but drops in performance for large mounting height differences. We attribute this to the fact that the trajectory is oftentimes not visible, when observed viewpoints at a largely different height. For the baselines, all the camera parameter mismatches consistently lead to performance drops. 
We note that even at small parameter mismatches, baseline performance falls below their 'matched'-camera results. This is partly because the 'cross' evaluation setting also applies rotational perturbations to the reference trajectory poses (see Sec.~\ref{subsec:sim_setup}), simulating realistic imperfect recordings.
\section{Model and Training Details}
\label{s_sec:impl_details}

This section provides implementation details for our model and training, see Sec.~\ref{sec:architecture} and Sec.~\ref{sec:training} of the main paper.

\subsection{Model Configuration}

Table~\ref{tab:model_config} summarises the key architectural parameters.
For all the transformer blocks, both multi-head attention and MLP follow a common pre-norm residual structure:
\[
\mathbf{x}
\;\leftarrow\;
\mathbf{x}
+ \;\text{LayerScale}\!\Big(\text{Op}\,\!\big(\text{RMSNorm}(\mathbf{x})\big)\Big),
\]
where $\text{Op}$ is either multi-head attention layer or MLP.
Drop-path regularization is applied to all residual branches during training, and per-head QK-norm (RMSNorm on queries and keys) is used in every attention layer.

\begin{table}[h]
\centering
\caption{Model architecture configuration}
\label{tab:model_config}
\begin{tabular}{ll}
\toprule
\textbf{Component} & \textbf{Configuration} \\
\midrule
Backbone & DINOv3 ViT-B/14 (frozen) \\
Input feature dimension & 768 \\
Hidden dimension $D$ & 256 \\
Attention heads $H$ & 8 \\
Head dimension $D/H$ & 32 \\
Trajectory encoder depth $L$ & 12\\
Query encoder depth & 6 ($L/2$) \\
Decoder depth & 6 ($L/2$) \\
Prediction head trunk depth & 3 \\
Prediction head iterations $K$ & 4 \\
FFN expansion ratio & 3 \\
Normalization & RMSNorm \\
Dropout / Att.\ dropout / Drop-path & 0.1\; /\; 0.1\; /\; 0.1 \\
RoPE head-dim split (spatial / temporal) & 24\; /\; 8 \\
RoPE base frequencies (spatial / temporal) & 500\; /\; 100 \\
\midrule
Total trainable parameters & $\sim$50M \\
\bottomrule
\end{tabular}
\vspace{-0.3cm}
\end{table}

\paragraph{Backbone and feature projection.}
A frozen DINOv3 ViT-B/14 backbone produces $14\!\times\!14 = 196$ patch tokens of
dimension 768 per input frame.
These are projected to the hidden dimension $D = 256$ via a linear layer
followed by GELU; separate projection layers are used for trajectory frames and
the query frame.
A learnable \textit{summary token} of dimension $D$ is prepended to each
frame's patch tokens, yielding $P = 197$ tokens per frame.
The camera token has two learned variants stored as a single parameter: one is
used for the query frame and the other is shared across all trajectory frames.

\paragraph{Rotary Position Embeddings.}
Each attention head's 32-dimensional space is partitioned into a
\textit{spatial} portion (24 dimensions) and a \textit{temporal} portion
(8 dimensions).
The spatial portion is further split equally into vertical and horizontal
halves, each encoded with 1D RoPE using the corresponding patch-grid coordinate
and base frequency 500.
The temporal portion receives 1D RoPE using the frame's sequential index within
the trajectory, with base frequency 100.

\paragraph{Trajectory encoder.}
The trajectory encoder stacks $L = 12$ \textit{Dual Attention} blocks.
Each block operates on tokens of shape $[B,\,S,\,P,\,D]$ and applies two
sub-layers in sequence:
\begin{enumerate}
  \item \textbf{Global self-attention.}
  Tokens are reshaped to $[B,\; S{\cdot}P,\; D]$ so that attention operates
  over all frames jointly, using full spatio-temporal RoPE.
  Followed by an MLP with expansion ratio 3.
\item \textbf{Spatial self-attention.}
  Tokens are reshaped to $[B{\cdot}S,\; P,\; D]$ so that attention operates
  independently within each frame, using spatial RoPE only.
  Followed by an MLP with expansion ratio 3.
\end{enumerate}

\paragraph{Query encoder.}
The query encoder has $L/2 = 6$ blocks, each consisting of a single spatial
self-attention sub-layer (spatial RoPE, followed by an MLP with expansion
ratio 3), identical in structure to the spatial sub-layer of the trajectory
encoder.
The output is passed through a feature adapter
(RMSNorm\,${\to}$\,Linear\,${\to}$\,GELU\,${\to}$\,RMSNorm)
before entering the decoder.

\paragraph{Decoder.}
The decoder has $L/2 = 6$ blocks, each interleaving cross-attention with a
local self-attention sub-layer:
\begin{enumerate}
  \item \textbf{Cross-attention.}
  Trajectory tokens (queries) attend to the adapted query tokens
  (keys and values), with spatial RoPE applied.
  Followed by an MLP with expansion ratio 3.

  \item \textbf{Local spatial self-attention.}
  A spatial self-attention sub-layer, identical in structure to those in the
  trajectory encoder, operates independently per frame.
  Followed by an MLP with expansion ratio 3.
\end{enumerate}
After all decoder blocks, the summary token (position 0) is extracted from each
frame, producing a $[B,\,S,\,D]$ tensor that is passed to the prediction head.

\paragraph{Iterative prediction head.}
The prediction head refines its output over $K = 4$ iterations using an
AdaLN-conditioned trunk inspired by DiT~\cite{peebles2023scalable}.
At each iteration $k$:
\begin{enumerate}
  \item The prediction from iteration $k{-}1$ is \textbf{detached} from the
  computation graph and embedded via a linear layer.
  At $k = 0$ a learned empty-pose token is used instead.

  \item The embedded vector is projected through SiLU\,${\to}$\,Linear to
  produce shift, scale, and gate vectors.
  These condition the input tokens via Adaptive Layer Normalization:
  \[
    \mathbf{x}'
    =
    \mathbf{x}
    + \;\text{gate}\;\odot\!\Big[
      (1 + \text{scale})\;\odot\;\text{RMSNorm}(\mathbf{x})
      + \text{shift}
    \Big].
  \]

  \item The conditioned tokens are processed by a trunk of 3 self-attention
  blocks, each followed by an MLP with expansion ratio 3.

  \item A two-layer MLP projects the output to a 4-dimensional residual
  update, which is accumulated onto the running prediction.
\end{enumerate}
The final predictions are mapped to bounded ranges: image coordinates
$\mathbf{p}_i = \tanh(\cdot) \in [-1,1]$;
a visibility logit $v_i$;
and a normalized distance
$d_i = \tfrac{1}{2}\big(\tanh(\cdot) + 1\big) \in [0,1]$.

\subsection{Training Configuration}

\paragraph{Optimization}
We employ the MUON~\cite{jordan2024muon} optimizer for all
two-dimensional weight matrices except those in the prediction head, while
AdamW optimizes the remaining parameters (biases, normalization layers, Layer Scale values, and all prediction-head parameters).
Both optimizers use Cautious Weight Decay~\cite{chen2025cautious} and share a cosine-annealing learning-rate schedule with linear warm-up.
Hyperparameters are listed below:
\begin{itemize}
  \item Base learning rate: $5\times10^{-4}$
  \item MUON momentum: $0.95$ (Nesterov)
  \item AdamW: $\beta_1 = 0.9$,\; $\beta_2 = 0.999$
  \item Weight decay: $\lambda = 0.05$
  \item Warm-up steps: 2\,000 
  \item Total training epochs: 40
  \item Gradient clipping: max norm $1.0$
\end{itemize}

\paragraph{Loss Configuration}
The total loss is a weighted sum of three terms, each evaluated over
non-padded sequence positions only:
\begin{itemize}
  \item $\lambda_{\text{pos}} = 10.0$\;: L1 loss on
  predicted image coordinates, computed only at frames where the target is
  visible.
  \item $\lambda_{\text{vis}} = 1.0$\;: Binary cross-entropy on visibility
  predictions.
  \item $\lambda_{\text{dist}} = 6.0$\;: L1 loss on
  predicted distances, normalised per trajectory by the maximum
  ground-truth distance, computed only at visible frames.
\end{itemize}
All three losses are aggregated across the $K = 4$ prediction-head iterations
with geometric weighting: iteration $k$ receives weight
$w_k = 0.8^{\,K-1-k}$, and the weighted sum is divided by $K$.
Because predictions are detached between iterations, each iteration contributes
an independent gradient.

\paragraph{Efficient Batching Strategy}
We exploit the asymmetric cost of the encoder--decoder architecture to
maximise throughput.
A batch consists of $N_\mathrm{T}$ trajectories whose features are encoded once
by $\mathcal{E}_\mathrm{T}$.
The encoded representations are then replicated and paired with up to
$8\!\times\! N_\mathrm{T}$ query views; the decoder processes all pairs
simultaneously, reusing the trajectory encodings.
Because trajectory encoding constitutes the dominant compute cost
(see Section~\ref{sec:architecture}), this amortization yields an effective
batch size of $8\!\times\! N_\mathrm{T}$ query images at relatively low additional cost.
In practice we set $N_\mathrm{T} = 8$, giving an effective batch size of 64
query images.

\paragraph{Implementation Details}
We represent variable-length trajectory sequences as PyTorch NestedTensors to
avoid padding overhead, and make use of \texttt{torch.compile} and PyTorch's BF16 mixed precision training together with
activation checkpointing.
All backbone features are pre-extracted before training begins.
Training takes approximately 4 days on a single NVIDIA RTX\,5090.

\paragraph{Scaling with Trajectory Length}
Online computation refers to the per-query-image computation performed
after the reference trajectory has been encoded. Its cost depends on the
number of active reference frames. Most trajectory-dependent operations
scale linearly with this number: in the query-trajectory fusion module,
cross-attention is applied from trajectory tokens to query tokens for each
reference frame, and the frame-wise attention blocks operate independently
within each reference frame. The prediction head is the only component with
quadratic scaling in the number of reference frames, as it applies
self-attention over one token per frame. Since this attention is
performed only over camera tokens rather than patch tokens, it remains
small for the trajectory lengths considered here. As shown in
Fig.~\ref{fig:scaling}, measured online inference time is approximately
linear up to 1000 reference frames, suggesting that the linear
trajectory-token operations dominate in this regime. In all reported
navigation experiments, we use trajectories subsampled to at most 40 frames.
\begin{figure}[h]
\centering
\definecolor{myblue}{HTML}{0173B2}
\begin{tikzpicture}
\begin{axis}[
    xlabel={Reference trajectory length (frames)},
    ylabel={Online latency (ms)},
    grid=both,
    width=1.0\columnwidth,
    height=0.3\textwidth,
    grid=major,
    grid style={dashed, gray!30},
    xmin=0,
    ymin=0,
    legend style={
        draw=none,
        fill=none,
        at={(0.02,0.98)},
        anchor=north west,
    },
    tick label style={font=\small},
    label style={font=\small},
]
\addplot+[
    myblue,
    thick,
    mark=*,
    mark size=1.7pt,
    error bars/.cd,
        y dir=both,
        y explicit,
]
table[
    x=seq_len,
    y=mean_ms,
    y error=stdev_ms,
] {
seq_len tokens mean_ms median_ms stdev_ms min_ms max_ms
5 985 3.451486 3.379296 0.199610 3.304256 4.771552
10 1970 3.378364 3.365584 0.049466 3.313152 3.637152
15 2955 3.389027 3.366656 0.071259 3.325472 3.737728
20 3940 3.366409 3.356320 0.048483 3.300288 3.623296
30 5910 3.320273 3.313376 0.034529 3.285696 3.522592
40 7880 3.359142 3.327632 0.110762 3.289312 4.093248
60 11820 3.498448 3.489952 0.064033 3.468192 4.048544
80 15760 4.056946 4.040624 0.102785 4.004288 4.877280
100 19700 4.677793 4.671040 0.042691 4.650400 5.001120
150 29550 6.963962 6.814848 0.400789 6.582656 8.800960
200 39400 8.381966 8.211664 0.322671 8.163104 9.706080
250 49250 10.237874 10.221584 0.156188 10.132672 11.506240
300 59100 12.060087 12.026736 0.146475 11.937216 13.232800
350 68950 14.314380 14.265840 0.192823 14.226336 15.621632
400 78800 16.103047 16.070736 0.151290 16.016672 17.340033
500 98500 19.376418 19.401616 0.796711 18.458719 21.535711
600 118200 23.346901 23.181664 0.920446 22.266081 25.767263
700 137900 26.328475 25.738432 0.992833 25.648705 29.574047
800 157600 30.551446 30.090672 0.749240 30.020960 32.912544
900 177300 34.284725 33.691216 0.962304 33.605473 37.959297
1000 197000 38.073791 37.800529 0.848797 37.270527 41.016640
};%
\addlegendentry{Mean latency $\pm$ 1 std. dev.}%
\end{axis}%
\end{tikzpicture}%
\caption{Online latency on an RTX 5090 over varying reference trajectory lengths over 100 runs for each setup.}
\vspace{-0.7cm}
\label{fig:scaling}
\end{figure}

\section{MASt3R and VGGT for Navigation}
\label{s_sec:vggt_study}
\begin{figure*}[t]
\centering
\def\svgwidth{\textwidth}
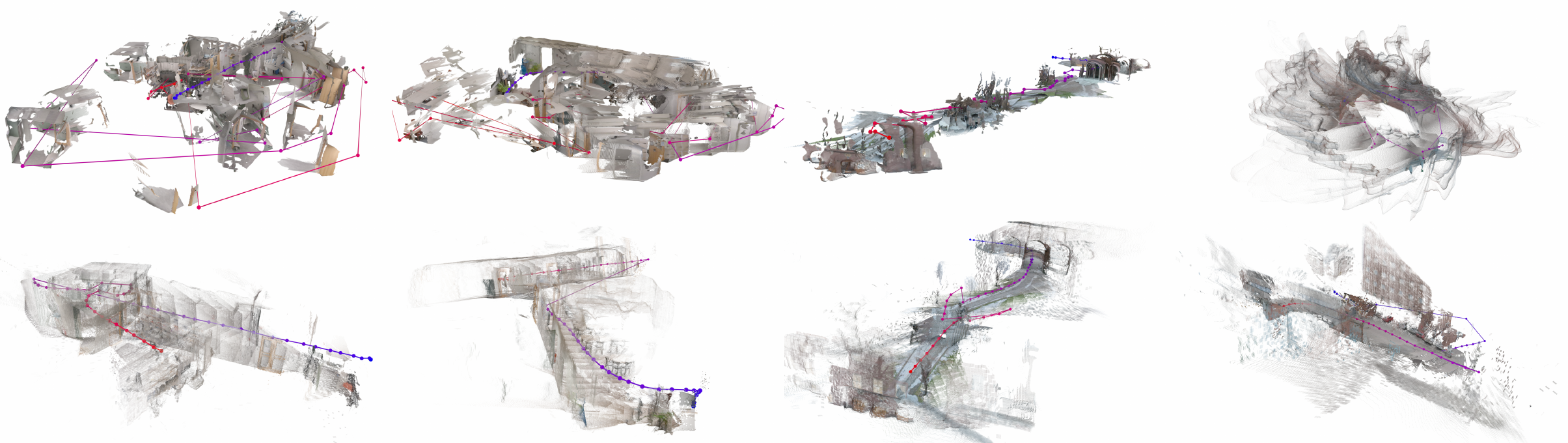
\caption{All real-world evaluation trajectories processed by VGGT (top row) and MASt3R (bottom row). Obtaining the full reconstruction took about $\sim 5\,\mathrm{min.}$ per scenario for MASt3R. For more accurate reconstructions, see Fig.~\ref{fig:realworld_setup} in the main paper.}
\vspace{-0.5cm}
\label{fig:mast3r_vggt}
\end{figure*}
In Section~\ref{sec:related_work} of the main paper, we discuss recent learned visual geometry methods including MASt3R~\cite{leroy2024mast3r} and VGGT~\cite{wang2025vggt}. Here, we evaluate learned geometry methods as potential navigation baselines by attempting to recover a usable relative trajectory representation from the reference and query images. In practice, MASt3R and VGGT frequently produce inconsistent poses or failed reconstructions on our real-world navigation trajectories, which prevents use without substantial additional heuristics. We therefore include qualitative failure cases and runtime measurements rather than reporting them as standard navigation baselines.

\subsection{Full Reconstruction}
We processed each trajectory using both methods following their official implementations (VGGT: Depthmap and Camera branch, MASt3R: full pairwise matching with global optimization). Fig.~\ref{fig:mast3r_vggt} shows the results. Both methods produce reconstructions that, for these trajectory lengths and environments, exhibit pose errors and inconsistencies that would make direct use for navigation challenging. Moreover, MASt3R required approximately $5\,\mathrm{min}$ per trajectory, and VGGT about 0.2$\,\mathrm{s}$.

\subsection{One-to-Many Matching Analysis}
\begin{figure}[h]
\centering
\def\svgwidth{0.7\columnwidth}
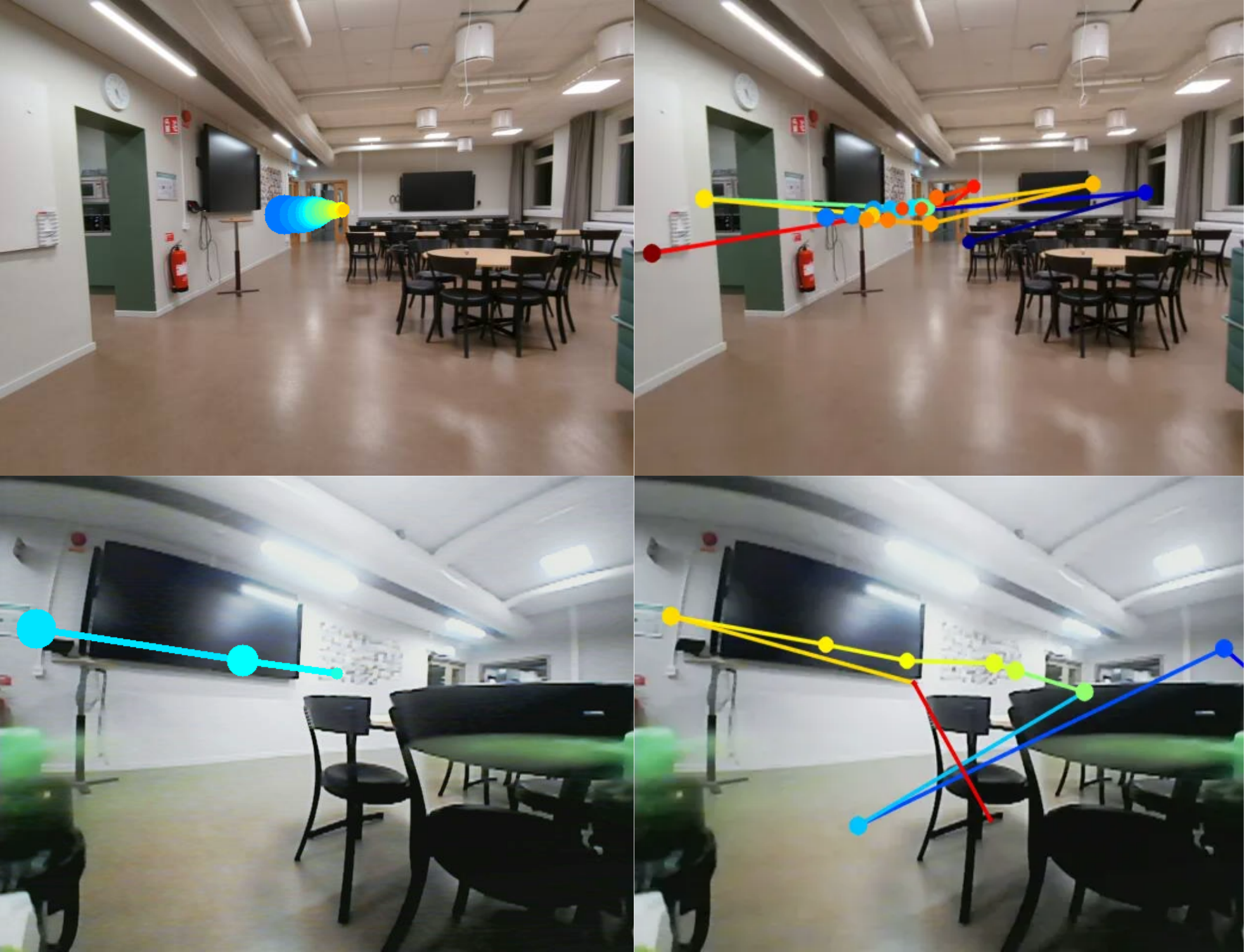
\caption{\methodname{}'s predictions (left) compared to MASt3R's one-to-many matching (right). We project MASt3R's predicted poses back to image space. The reference trajectory can be found on the project page (Indoors 2). We show the results for on-trajectory views (top) and off-trajectory views from on-board the quadcopter (bottom).}
\vspace{-0.5cm}
\label{fig:ours_mast3r}
\end{figure}
We also tested MASt3R's one-to-many matching mode, which matches all frames against a single query image, which is closer to our online setting. Fig.~\ref{fig:ours_mast3r} compares the resulting pose predictions (projected to image space) against \methodname{}'s predictions for both on-trajectory and off-trajectory query views. MASt3R's predictions are noisy and inconsistent, while \methodname{} produces stable image-space outputs suitable for control.

These results suggest that the methods do not directly provide meaningful real-time navigation guidance, at least on trajectories similar to ours.
\section{Controller Implementation Details}
\label{s_sec:contr_details}
This section details the controllers discussed in Sec.~\ref{sec:control} in the main paper.
\subsection{Yaw Controller with constant forward velocity}
From the current predictions $(\mathbf{p}_{1:N}, v_{1:N}, d_{i, N})$, we first extract the indices corresponding to \textit{visible} predicted points, $\mathcal{I}_{\mathrm{vis}}$, by thresholding the visibility confidence:
\begin{equation}
    \mathcal{I}_{\mathrm{vis}} = \{ i \mid \sigma(v_i) > 0.5, i \in [1, N] \}.
\end{equation}
Since prediction indices naturally follow the trajectory ordering, this sequence is sorted, where the higher indices correspond to points closer to the end of the trajectory. From this sequence, we identify the current \textit{reference trajectory index} $k \in \mathcal{I}_{\mathrm{vis}}$ as the visible point closest to the robot:
\begin{equation}
    k = \operatorname*{argmin}_{i \in \mathcal{I}_{\mathrm{vis}}} d_i.
\end{equation}
The traversal direction $s \in \{+1, -1\}$ is determined by comparing this trajectory index $k$ to the goal index: $s = \mathrm{sgn}(g - k)$. We apply a small lookahead $\Delta$ within the \textit{visible sequence space} to determine the target point. Let $j$ be the position of $k$ in the sequence of visible indices (i.e., $\mathcal{I}_{\mathrm{vis}}[j] = k$). The target point $\mathbf{p}_{\mathrm{target}}$ is selected as:
\begin{equation}
    \mathbf{p}_{\mathrm{target}} = \mathbf{p}_{m}, \quad \mathrm{where } ~m = \mathcal{I}_{\mathrm{vis}}[\mathrm{clip}(j + s \cdot \Delta, 1, |\mathcal{I}_{\mathrm{vis}}|)].
\end{equation}

For views in which the trajectory is out-of-view (e.g. below the camera, or in sharp turns), we found that our model still provides meaningful guidance by predicting points at the border of the frame in the direction of the out-of-view trajectory, but (correctly) predicts them as not visible. To take advantage of that, we consider close, non-visible points as target points if $\mathcal{I}_\mathrm{vis} = \emptyset$.
Finally, we apply a P-controller to minimize the horizontal pixel error between the image center and $\mathbf{p}_{\mathrm{target}}$, while commanding constant forward velocity.

\subsection{Model Predictive Path Integral Control}
MPPI is a control method to solve stochastic Optimal Control Problems for discrete-time dynamical systems
\begin{align*}
    \mathbf{x}_{k+1} = \mathbf{F}(\mathbf{x}_k, \mathbf{v}_k), \hspace{0.5cm} \mathbf{v}_k \sim \mathcal{N}(\mathbf{u}_k, \mathbf{\Sigma}).
\end{align*}
MPPI samples $M$ random control input sequences $\mathbf{v}_{0:K-1}^{(1:M)}$ of length $K-1$ and forward simulates the system dynamics given the current state $\mathbf{x}_0$ to obtain $X^{(m)} = [\mathbf{x}_0, \mathbf{F}(\mathbf{x}_0, \mathbf{v}_0^{(m)}), \dots, \mathbf{F}(\mathbf{x}_{K-1}, \mathbf{v}_{K-1}^{(m)})]$. In our implementation, we use a simplified single integrator model which independently controls the linear velocity and yaw rate of the quadcopter, i.e.
\begin{align}
    \mathbf{x}_{k+1} = \mathbf{x}_k + \delta t \cdot \mathbf{u}_k
\end{align}
with state $\mathbf{x} = [p_x, p_y, \psi]^T$ and control input $\mathbf{u}=[v_x,v_y, \omega_{\psi}]^T$. The height $p_z$ is independently controlled by keeping the vertical position of the goal point in the image-space center. 

Then, given the state rollouts and a cost function $J(X)$ to be minimized, each rollout is weighted by an importance sampling weight 
\begin{align}
    w^{(m)} = \frac{1}{\eta} \exp \left(- \frac{1}{\beta} \left(J(X^{(m)}) - \rho)\right)\right),\label{eq:ImportanceWeight}
\end{align}
where $\eta$ is a normalization constant ensuring $\sum_{m=1}^M w^{(m)} = 1$, $\rho = \min_m J(X^{(m)})$ is subtracted for numerical stability and $\beta$ is the \emph{inverse temperature} which serves as a tuning parameter for the sharpness of the control distribution. Finally, an approximate optimal control sequence can be obtained as
\begin{align*}
    \mathbf{u}_{0:K-1}^* = \sum_{m=0}^M w^{(m)} \mathbf{v}_{0:K-1}^{(m)},
\end{align*}
which is a weighted average of sampled control trajectories and is applied in a receding horizon fashion. For a detailed discussion and theoretical properties, we refer to \cite{williams2017model}.
\subsection{Cost Terms}
Our cost function
\begin{align}
    J(X) = \sum_{i=1}^K w_\mathrm{goal}\mathcal{C}_\mathrm{goal} + w_\mathrm{vis} \mathcal{C}_\mathrm{vis} + w_\mathrm{col} \mathcal{C}_\mathrm{col}\label{eq:mppi_cost}
\end{align}
consists of three terms that penalize deviation from the goal point, losing visibility of the goal point and collisions in the environment.

The first term our cost formulation in Eq.~\eqref{eq:mppi_cost} rewards progress along the path by penalizing distance to a goal point. Since only visible image-space predictions $\mathcal{I}_\mathrm{vis}$ are available, we first ground them in 3D to obtain a reference trajectory. We use UniDepthV2~\cite{piccinelli2025unidepthv2} to predict depth and camera parameters from the current query image $I_q$. Given the model predictions $(\mathbf{p}_i, v_i, d_i)_{i \in \mathcal{I}_{\text{vis}}}$, we project the unscaled image-space points into the depth image using the normalized distances $d_i$. As metric scale is unknown, we scale the resulting trajectory as far as possible within the collision-free space and use the scaled points $\tilde{\mathbf{p}}_1^{\text{ref}}, \tilde{\mathbf{p}}_2^{\text{ref}}, \dots$ as the reference trajectory.

The goal cost is implemented as a simple two-norm, i.e.
\begin{align}
    \mathcal{C}_{\text{goal}} = \lVert [p_x, p_y]^T - \mathbf{g}\rVert
\end{align}
where $\mathbf{g} \in \{\tilde{\mathbf{p}}_1^{\text{ref}}, \tilde{\mathbf{p}}_2^{\text{ref}}, \dots\}$ can be any point on the reference trajectory depending on the strategy. For instance, the goal point can be chosen as the furthest progressed one to encourage short-cutting behavior, or an earlier point on the path to prefer trajectory tracking. In practice, we use the third point to balance the two behaviors.


For collision avoidance, we use the predicted depth image to create a distance field $DF: p_x \times p_y \mapsto \mathbb{R}_{\geq 0}$ and add a penetration cost as
\begin{align}
    \mathcal{C}_{\text{coll}}= \mathds{1}_{DF(p_x, p_y) \leq r} \cdot DF(p_x, p_y)
\end{align}
where $r$ denotes the collision radius of the quadcopter.

Lastly, in order to keep the predictions in the FOV, we add an additional cost term
\begin{align}
    \mathcal{C}_{\text{vis}} = \left(\mathrm{atan2}\left(g_y-p_y, g_x - p_x\right) - \psi\right)^2
\end{align}
that penalizes paths that lose track of the predicted points. Overall, the first two cost terms address collision-free path tracking through velocity commands $v_x, v_y$ while $\mathcal{C}_{\text{vis}}$ encourages the robot to separately make use of $\omega_{\psi}$ to keep the goal point in the FOV. Throughout our experiments, we use weights $w_{\text{goal}}=10, w_{\text{coll}}=100$ and $w_{\text{vis}} = 10$.
\end{document}